\documentclass[lettersize,journal]{IEEEtran}
\IEEEoverridecommandlockouts                              % This command is only needed if 
\usepackage{epsfig} % for postscript graphics files
\usepackage{amsmath} % assumes amsmath package installed
\usepackage{amssymb}  % assumes amsmath package installed

\usepackage{cite}[nocompress]

\usepackage{bm}
\usepackage{caption}
\usepackage{subcaption}
\usepackage[hidelinks]{hyperref}
\newcommand{\va}{\mathbf a}

\newcommand{\vf}{\mathbf f}

\newcommand{\vv}{\mathbf v}

\title{\LARGE \bf
	Impact-Aware Control using Time-Invariant Reference Spreading%*
}

\author{Jari van Steen%$^{1}$% <-this % stops a space
}

\author{Jari van Steen, Nathan van de Wouw and Alessandro Saccon%$^{1}$% <-this % stops a space
 \thanks{This work was partially supported by the Research Project I.AM. through the European Union H2020 program under GA 871899.}
 \thanks{Jari van Steen, Nathan van de Wouw and Alessandro Saccon are with the Faculty of Mechanical Engineering, Eindhoven University of Technology, 5612 AE Eindhoven, The Netherlands (e-mail: j.j.v.steen@tue.nl, n.v.d.wouw@tue.nl, a.saccon@tue.nl)}
}

\begin{document}

	\maketitle
	\thispagestyle{empty}
	\pagestyle{empty}

    \begin{abstract}
    With the goal of increasing the speed and efficiency in robotic manipulation, a control approach is presented that aims to utilize intentional simultaneous impacts to its advantage. 
	This approach exploits the concept of the time-invariant reference spreading framework, in which partly-overlapping ante- and post-impact reference vector fields are used. These vector fields are coupled via an impact model in proximity of the expected impact area, minimizing the otherwise large impact-induced velocity errors and control efforts. We show how a nonsmooth physics engine can be used to construct this impact model for complex scenarios, which warrants applicability to a large range of possible impact states without requiring contact stiffness and damping parameters. 
	In addition, a novel interim-impact control mode provides robustness in the execution against the inevitable lack of exact impact simultaneity and the corresponding unreliable velocity error during the time when contact is only partially established. This interim mode uses a position feedback signal that is derived from the ante-impact velocity reference to promote contact completion, and smoothly transitions into the post-impact mode. 
	An experimental validation of time-invariant reference spreading control is presented for the first time through a set of 600 robotic hit-and-push and dual-arm grabbing experiments.
\end{abstract}

\section{Introduction}\label{sec:introduction}
  
    % NOTE: Much of this is currently copied from the IFAC WC paper, assuming it is an "evolved paper"
    Exploiting impacts in robotics has been a topic of interest for many years, enabling complex robotic locomotion tasks such as running \cite{Katz2019, Wensing2013} and jumping \cite{Zhang2020}. 
    Recently, research interest has also started to increase towards robotic impact-aware manipulation tasks such as hitting \cite{Khurana2021}, catching \cite{Yan2024} or fast grabbing \cite{Bombile2022,Dehio2022,Zermane2024} of objects. One application domain that can benefit from this research is the field of material handling in logistics processes. Fast and reliable automated solutions with a small footprint are sought in this domain, but at the moment only human operators are capable of performing these tasks, which are often injury-prone and not fully ergonomic. %Impact-aware manipulation is a strategy to enable fast grab-and-toss operations, besides the more classical pick-and-place.   
    %, where more efficient automated solutions are desired for tasks where humans currently excel in speed and reliability due to their ability to utilize impacts, such as rapid pick-and-place operations. 
    Within this context, impact-aware dual-arm manipulation could provide a solution for fast material handling of heavy objects, potentially allowing for human-like swift pick-and-place operations without the need of creating a custom end effector for different types of handled products \cite{Smith2012,Benali2018}. 
    
    Utilization of impacts in robotic manipulation in a human-like fashion can decrease cycle times, but requires addressing multiple challenges. Firstly, it needs to be ensured that hardware does not get damaged due to potentially large peak forces on the robots induced by impacts \cite{Dehio2022,Ostyn2021}. 
    But secondly, even when impacts stay within safe limits, the rapid velocity transitions that result from the impact can result in a peak in the error between the desired and actual velocity of the robot under traditional tracking control as a result of the inevitable mismatch between the actual and the predicted impact times \cite{Biemond2013,Leine2008}. This, in turn, results in undesired peaks in the actuator commands, potentially inducing vibrations, controller destabilization, hardware damage and increased energy consumption. 

    \begin{figure}
		\centering
		\includegraphics[width=\linewidth]{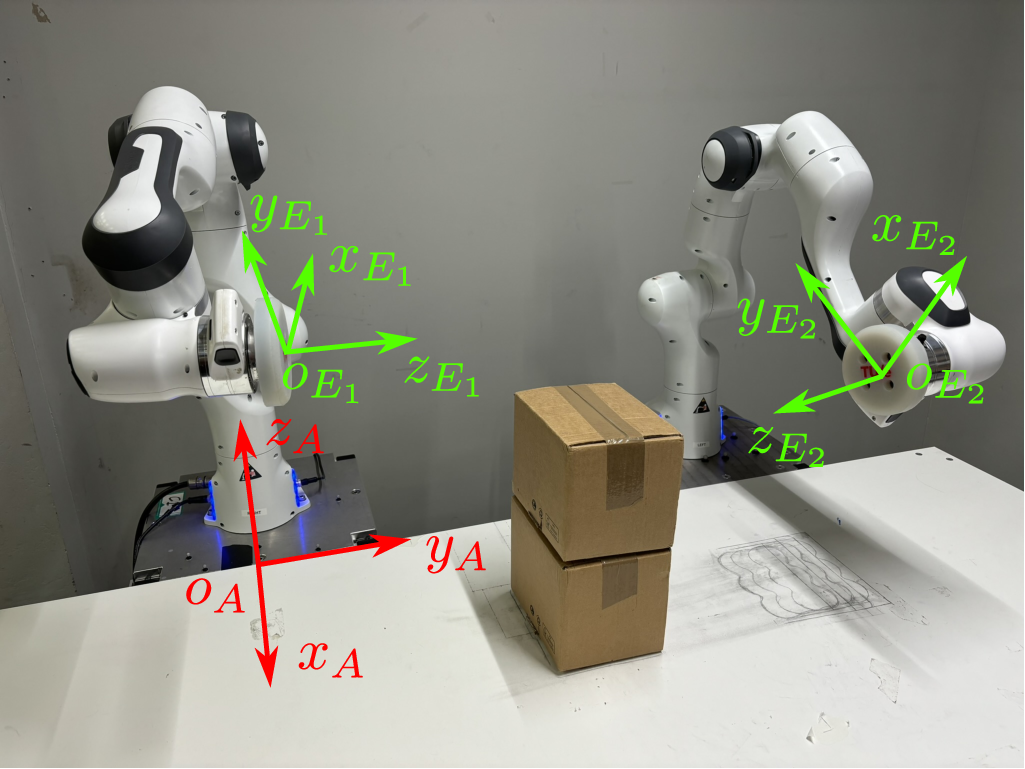}
		\caption{Depiction of the dual-arm robotic setup used for the experimental validation of the control approach presented in this work.}
		\label{fig:setup}
	\end{figure}	    
 
    This work focuses on the extension of a framework for robotic manipulation under \textit{nominally simultaneous impacts} \cite{Rijnen2019} suitable for single- and dual-arm manipulation. In a nominally simultaneous impact, contact between the environment and the robot or robots in all relevant contact areas is ideally established simultaneously with nonzero velocity. In the development of the proposed control strategy, impacts are modeled as instantaneous events, following theory of nonsmooth mechanics \cite{Brogliato2016, Glocker2006}.
    It is, however, inevitable that impacting surfaces are not perfectly aligned at the time of impact, or that one robot establishes contact just before the other robot (in dual-arm manipulation), for example, due to uncertainties in the environment. 
    As a result, a series of unplanned intermediate impacts and a corresponding unpredictable series of velocity jumps will typically occur, where it is generally not possible to estimate the contact state. As confirmed by simulations and experiments, this implies that velocity feedback control cannot be reliably used during an impact sequence \cite{Steen2024,Yang2021}. The control approach proposed in this work combines the ability to handle simultaneous impacts with a time-invariant framework for generating references, and is called \emph{time-invariant reference spreading}. In this work, we present for the first time an application of this time-invariant reference spreading framework on a real-world robotic setup, as well as a thorough experimental validation. 
    In the remainder of this section, we will present the related works in more detail, followed by the precise research contribution and an outline of this work.

    \subsection{Related works}
    
    In recent years, a handful of dedicated control methods have been developed to tackle the execution of motions that include impacts while avoiding unwanted spikes in the control inputs. As an example, \cite{Biemond2013,Biemond2016,Baumann2018} defines the tracking error through a distance function that includes knowledge on the predicted velocity jump, removing the velocity tracking error peak caused by a mismatch between the planned and actual impact timing. Similarly, \cite{Forni2011}  performs tracking control of a bouncing ball using an additional mirrored reference to address a possible impact timing mismatch, in this case assuming a fully elastic impact. 
    An explicit transition phase is defined in \cite{Morarescu2010} to perform tracking control during contact transitions with chattering caused by impacts. 
    Alternatively, \cite{Yang2021} projects the velocity tracking error onto a subspace that is invariant to the impact event, temporarily removing velocity feedback for the control objectives affected by the impact, but retaining full control over all remaining control objectives. 
 
    The removal of unwanted input spikes caused by impacts has also been addressed by the framework of \emph{reference spreading} (RS) \cite{Rijnen2019,Steen2022,Steen2024}. RS, as introduced in \cite{Saccon2014} and expanded in \cite{Rijnen2015,Rijnen2017,Rijnen2019a,Rijnen2020} and references therein for single impact scenarios, is a hybrid tracking control approach that deals with impacts by defining ante- and post-impact references as a function of time that are coupled by an impact map at the nominal impact time, and overlap about this time. 
    It is ensured that the reference corresponds with the actual contact state of the robot by switching the reference based on impact detection, which avoids error peaking and related spikes in the control inputs. In \cite{Rijnen2019}, a strategy to deal with simultaneous impacts is proposed, defining an interim-impact mode that uses only feedforward signals as long as contact is only partially established. This framework is extended in \cite{Steen2022} by also using position feedback during the interim mode to pursue persistent contact establishment without relying on velocity feedback. 
    While \cite{Rijnen2019} and \cite{Steen2022} only included a numerical validation based on simulations, an experimental validation of RS for simultaneous impacts is provided in \cite{Steen2024} using a dual-arm robotic setup targeted to perform a quick grabbing maneuver. %This work furthermore provides a further updated interim mode formulation. 
    Furthermore, the approaches in \cite{Steen2022,Steen2024} are cast into the quadratic programming (QP) robot control framework \cite{Bouyarmane2019,Salini2010}, similar to other impact-aware QP control approaches like \cite{Dehio2021, Wang2019}. This allows to include constraints ensuring for example collision avoidance and adherence to joint limits away from impacts, which are essential in real applications. 

	While \cite{Rijnen2019,Steen2022,Yang2021,Morarescu2010} use traditional \textit{time-based} references, \cite{Steen2022a} has introduced a \textit{time-invariant} version of RS, which has been extended in \cite{Steen2022b} for dual-arm manipulation, and will be further extended in this work as highlighted in Section \ref{sec:contribution}. As opposed to time-based tracking, the time-invariant nature of this approach enjoys the positive features of path following \cite{Aguiar2005}, maneuver regulation \cite{Hauser1995} and dynamical systems \cite{Billard2022} approaches, preventing the systems to unnecessarily accelerate or decelerate in the presence of disturbances and deviations caused by temporary collision avoidance or conflicting tasks. 
	The approach also suits a large range of initial conditions without requiring replanning of the reference due to the absence of a predefined path that is to be followed as with traditional tracking control. 
	The references in \cite{Steen2022a,Steen2022b} are prescribed by desired ante- and post-impact vector fields, which overlap in position around the surface where the impact is expected to occur, to make sure a reference corresponding to the contact state is also at hand when the impact occurs away from this expected surface. At this surface, the ante- and post-impact velocity fields are coupled by an impact map \cite{Glocker2006,Brogliato2016}. %The interim mode philosophy of \cite{Steen2022} is followed by constructing
	In \cite{Steen2022a,Steen2022b}, the interim mode control developed in \cite{Steen2022} is extended to the case of time-invariant reference vector fields, constructing a position feedback signal through time integration of the ante-impact velocity reference. To promote simultaneity between the impacts for dual-arm manipulation when using with time-invariant references, \cite{Steen2022b} also introduces a synchronization approach between the two arms, resulting in a near-simultaneous impact regardless of the initial condition of the robots.  
	Recent approaches, like \cite{Khurana2021} and \cite{Bombile2022}, focusing on hitting and dual-arm grabbing of objects, respectively, also employ such a time-invariant approach in the context of impact-aware manipulation and originate from the well-known framework of dynamical systems-based robot control \cite{Billard2022,Salehian2018}. These approaches, however, focus on motion generation and not on control challenges such as the removal of impact-induced input peaks, and are thus complementary to the present work. While a numerical validation of time-invariant RS using simulations is provided in \cite{Steen2022a,Steen2022b}, no experimental validation has yet been performed. 

    As mentioned before, both time-based and time-invariant RS rely on an impact map to ensure that the velocity jump embedded in the ante- and post-impact references matches that of the real system. This impact map can be determined in different ways. For systems involving only a few degrees of freedom and a few contact points, an analytical impact map can be determined, which forms a closed-form expression to determine the post-impact velocity as a function of the ante-impact configuration and velocity. This method has been used in \cite{Saccon2014,Rijnen2019,Rijnen2019a,Rijnen2020,Steen2022} while developing and validating the RS approach. However, such an approach is not scalable to complex and realistic scenarios that could emerge in real applications. Alternatively, \cite{Steen2024} extracts the impact map from physical experiments. This removes the need of computing the impact map, but is also not scalable to situations where impacts are expected to happen with previously unhandled objects or in different configurations, potentially unknown a priori.
    
    An approach to extract the impact map from a set of simulations was proposed in \cite{Steen2022b}, using it for a time-invariant RS approach and validating the approach through a different set of simulations. The potential of using a physics engine based on nonsmooth mechanics to compute a rigid impact map has been confirmed in \cite{Steen2024a} by validating this computed impact map against a set of physical experiments. The validation in \cite{Steen2024a} showed only a 3.55\% average mismatch between the predicted velocity jump and the experimentally identified velocity jump for a set of dual-arm grabbing and hit-and-push motions, which is deemed sufficiently low to be reliably applied within the RS framework. The experimental validation of time-invariant RS presented in this work will therefore use the simulation-based approach developed in \cite{Steen2024a} to compute the rigid impact map.

    %A common rigid impact model, for example used in \cite{Kirner2024,Proper2023,Yang2021}, is the analytical impact model, which maps the ante-impact configuration and velocity into a post-impact velocity through a closed-form expression, while assuming the post-impact configuration remains equal due to the infinitesimal impact duration. However, for complex manipulation tasks, such as dual-arm grabbing with two 7 degree of freedom (DOF) robotic arms, an analytical impact model cannot be easily derived. 
    %An alternative is presented in \cite{Steen2022b}, which computes the rigid impact map by simulating the impact event using a compliant contact model. Such a compliant contact model can, however, be computationally expensive due to the small timestep required for accurate results, and demands challenging estimation of stiffness and damping contact parameters. In \cite{Steen2024}, the rigid impact map is extracted from physical experiments, requiring a training procedure and thus preventing the manipulation of previously unhandled objects. 

    %%%% Copied and unaltered

    \subsection{Contribution}\label{sec:contribution}

    The main contribution of this paper is the formulation and experimental validation of a control framework to perform robotic motions that contain nominally simultaneous impacts, such as dual-arm grabbing. Our aim is to remove control input peaks  otherwise caused by impact-induced velocity jumps, while accurately tracking the time-invariant velocity vector fields that prescribe the motions. 
    This paper builds on preliminary work on the framework of time-invariant RS in \cite{Steen2022a,Steen2022b} and is a direct extension of \cite{Steen2022b}. Time-invariant RS    uses overlapping ante- and post-impact velocity reference fields that are coupled via the impact dynamics, together with a suitable control structure with an ante-impact, interim and post-impact control mode. 
    
    This work shows for the first time an experimental validation of time-invariant RS, demonstrating our approach using a robotic dual 7 degree of freedom (DOF) arm setup for a hit-and-push and a dual-arm grabbing use cases. While \cite{Steen2022a} and \cite{Steen2022b} did include a numerical validation using simulations on a planar use cases, no experimental validation was performed in these works. We also demonstrate for the first time that we can reliably use an impact map obtained through a nonsmooth physics engine in this time-invariant RS approach to couple ante- and post-impact velocity reference fields, building on the work of \cite{Steen2024a} that experimentally validated the use of nonsmooth simulations to build an impact map. The approach of \cite{Steen2022b} did present a similar simulation-based approach using Matlab simulations to evaluate the impact map, which lacked the scalability of simulations using an established physics engine, while the smooth contact model that was used requires knowledge on contact stiffness and damping parameters that are difficult to obtain. 
    Lastly, the control approach from \cite{Steen2022a,Steen2022b} is modified to be effective in a real-life setting, in particular through the definition of a novel interim-impact mode, which is active when contact is only partially established. The interim-impact mode defined in \cite{Steen2022a,Steen2022b} uses knowledge about the moment when the impact sequence is finished, which is not available in an experimental setting. This work therefore defines a novel interim mode that builds on ideas of \cite{Steen2022a,Steen2022b} by generating a position feedback signal based on the velocity field references, but is also effective in an experimental setting.
    
    %bundles ideas from the experimentally validated interim mode presented in \cite{Steen2024} for time-based RS with the philosophy from the interim-impact mode in \cite{Steen2022a,Steen2022b} to generate a position feedback signal based on the velocity field references.

    \subsection{Outline}
 
	The outline of this paper is as follows. In Section \ref{sec:eom}, we provide equations of motion of the dual-arm robot system used to demonstrate the proposed control approach. 
	Section \ref{sec:reference} presents the approach to formulate the ante- and post-impact reference velocity fields, coupled via an impact map through simulations with a nonsmooth physics engine. Section \ref{sec:control} then describes the control framework consisting of the ante-impact, interim, and post-impact modes, used to track these references. 
	In Section \ref{sec:validation}, we show an experimental validation of the proposed framework and a comparison with three baseline approaches for a hit-and-push and a dual-arm grabbing use case, before drawing conclusions in Section \ref{sec:conclusion}.

\section{Equations of Motion}
\label{sec:eom}

    While the control framework described in this work is applicable to different robots and tasks containing single or ideally simultaneous intentional impacts, we will illustrate the approach on the dual 7DOF robot-arm setup depicted in Figure \ref{fig:setup} for a hit-and-push and a dual-arm grabbing task. A silicone end effector, with a frame $E_i$ for robot index $i\in \{1,2\}$, is attached to both robots. Using a shortened variant of the notation of \cite{Traversaro2019}, we denote the position and rotation of the end effector frames $E_i$ with respect to the inertial frame $A$ as $\bm p_i := {}^A \bm p_{E_i}$ and $\bm R_i := {}^A \bm R_{E_i}$, respectively. We denote its twist as $\vv_i = \left[\bm v_{i}, \bm \omega_i\right]  := {}^{E_i[A]} \vv_{A,E_i}$, and its acceleration as $\va_i = \left[\bm a_{i}, \bm \alpha_i\right]  := {}^{E_i[A]} \va_{A,E_i}$ with linear and angular velocity, and linear and angular accelerations $\bm v_{i}, \bm \omega_{i}, \bm a_{i}, \bm \alpha_{i}$, respectively.

    Each robot contains 7 actuated joints with joint displacements $\bm q_i \in \mathbb{R}^7$, where the drivetrain is assumed to be rigid. The end effector twists and accelerations can be expressed in terms of the joint velocities and accelerations as
	\begin{equation}
	\vv_i = \bm J_i(\bm q_i) \dot{\bm{q}}_i,
	\end{equation}
        \begin{equation}
	\va_i = \bm J_i(\bm q_i) \ddot{\bm{q}}_i + \dot{\bm J}_i(\bm q_i, \dot{\bm q}_i) \dot{\bm{q}}_i 
	\end{equation}
	with geometric Jacobian $\bm J_i(\bm q) :=  {}^{E_i[A]}\bm J_{A,E_i}(\bm q)$.     
    The equations of motion of both robots have the form 
	\begin{equation}\label{eq:eom}
	\left(\bm M_i(\bm q_i) + \bm B_{\theta,i}\right)\ddot{\bm q}_i + \bm h_i(\bm q_i,\dot{\bm q}_i)= \bm \tau_i + \bm J_{i}^T(\bm q_i)  \mathbf{f}_i % {}^{i[A]} \bm J_{A,i}^T {}_A \vf_i
	\end{equation} 
	with mass matrix $\bm M_i(\bm q_i)$, vector of gravity, centrifugal and Coriolis terms $\bm h_i(\bm q_i,\dot{\bm q}_i)$, commanded joint torques $\bm \tau_i$, and contact wrench $\vf_i$. 
    The term $\bm B_{\theta,i}$ represents the desired apparent motor inertia, which is a parameter appearing in the low-level torque control law applied to the Franka Robotics arms used in this work. As highlighted in \cite{Ott2015,Arias2024,Steen2024a}, we can show that \eqref{eq:eom}, which includes $\bm B_{\theta,i}$, is the most accurate representation of the equations of motion for a torque controlled robot with flexible joints such as the Franka Robotics arms, given the assumption of a rigid joint transmission with commanded (instead of measured) joint torques $\bm \tau_i$. In \cite{Steen2024a}, this is further backed by an experimental validation of the impact map through simulations with a nonsmooth physics engine. In this validation, the impact map was found to be most accurate for a robot model with motor inertia $\bm B_{\theta,i}$, rather than a robot model with no motor inertia or classical link-side motor inertia obtained by scaling the motor inertia with the gear ratio. 
    For ease of notation, the explicit dependency of the various terms appearing in \eqref{eq:eom} on $\bm q_i$ (or $\dot{\bm q}_i$) is dropped for the remainder of the paper.

\section{Time-invariant reference formulation}
\label{sec:reference}

    In this section, the framework used to formulate time-invariant velocity references for impact tasks is presented. In line with other approaches that use time-invariant references for impact tasks such as \cite{Bombile2022}, we define two separate velocity fields; an ante-impact and a post-impact reference. First, the formulation of the ante-impact reference velocity field is specified, followed by the formulation of a post-impact reference field, which is coupled with the ante-impact reference through an impact map generated using simulations with a nonsmooth physics engine.

    %as well as a method to formulate a coupled velocity reference for the box in the post-impact mode. 

\subsection{Ante-impact velocity field}\label{sec:ante_ref_formulation}

\begin{figure}
		\centering
		\begin{subfigure}[b]{\linewidth}
			\centering
			\includegraphics[width=\textwidth]{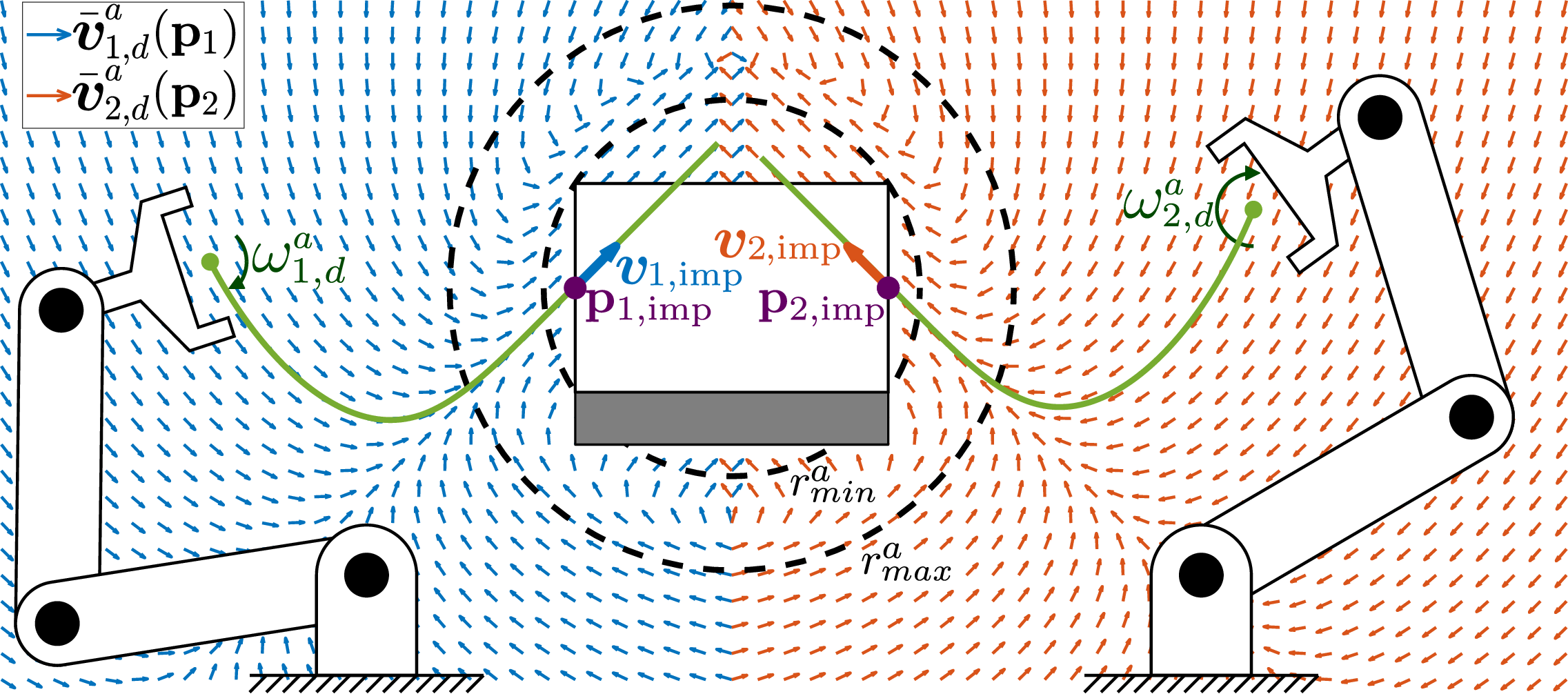}
			\caption{Ante-impact velocity reference with nominal end effector path.}
			\label{fig:ante_reference}
		\end{subfigure}
		\begin{subfigure}[b]{0.55\linewidth}
			\centering
			\includegraphics[width=\textwidth]{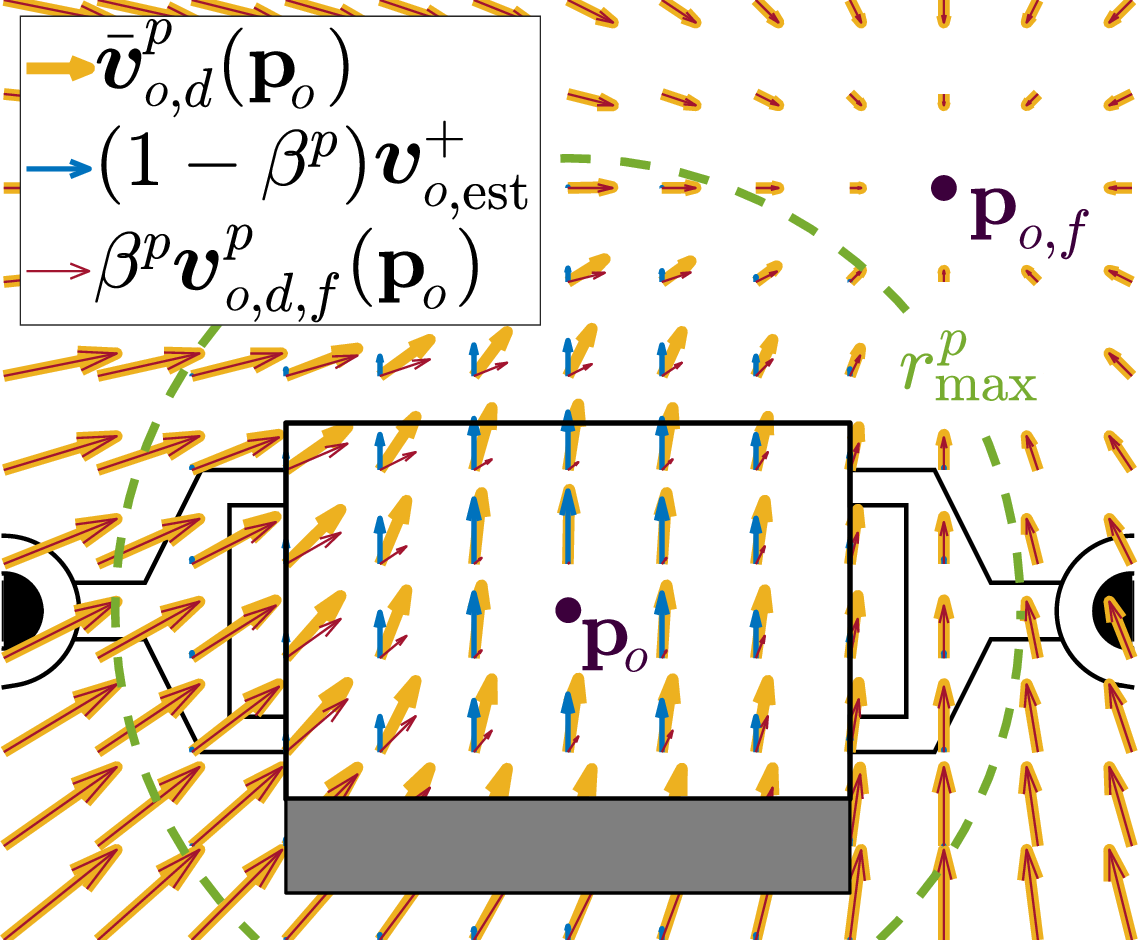}
			\caption{Post-impact velocity reference}
			\label{fig:post_field}
		\end{subfigure}
		\hfill
		\begin{subfigure}[b]{0.34\linewidth}
			\centering
			\includegraphics[width=\textwidth]{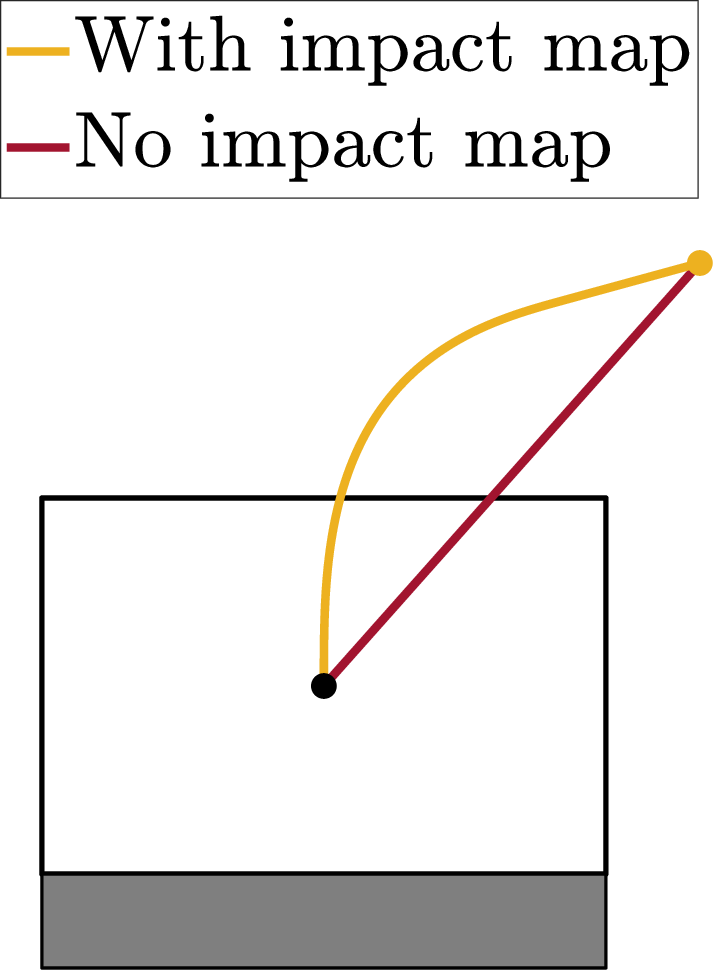}
			\caption{Nominal box path.}
			\label{fig:post_path}
		\end{subfigure}
		\caption{Ante- and post-impact velocity reference vector fields depicted for a dual-arm 3DOF planar use case: (a) shows the ante-impact velocity reference field for robot 1 (blue) and robot 2 (red) together with the nominal end effector path (green);
        (b) shows the post-impact velocity reference field with $r_\text{min}^p = 0$ (yellow) and the two velocity fields that together comprise this reference field (blue and red). The blue field represents the term that corresponds to the predicted post-impact velocity $\bm v^+_{o,\text{est}}$, and the red field represents the term that eventually ensures convergence to the point $\bm p_{o,f}$; (c) shows the nominal path followed by the center of the object when the velocity reference field is followed in yellow, versus the red path taken when the red vector field from (b) is followed without accounting for the predicted post-impact velocity.}
		\label{fig:post_reference_complete}
	\end{figure}

    The aim of the ante-impact velocity reference is to generate a vector field that guides the end effector of each robot to a given impact state. This reference velocity field is applicable to both the hit-and-push and the dual-arm grabbing use cases for which the proposed approach is demonstrated. %, is mostly identical to the velocity field formulated in \cite{Steen2022b}, with the angular velocity reference adjusted to suit a three-dimensional use case rather than the planar one used in \cite{Steen2022b}. 
    To start, a desired impact location $\bm p_{i,\text{imp}}$ is defined together with a desired impact velocity $\bm v_{i,\text{imp}}$. As shown in Figure \ref{fig:ante_reference}, $\bm p_{i,\text{imp}}$ and $\bm v_{i,\text{imp}}$ represent intuitive parameters used to create the reference vector field. %, following approaches like \cite{Bombile2022}. 
    An ante-impact velocity field $\bm v^a_{i,d}(\bm p_i)$ is defined as a first step, given by
    \begin{equation}\label{eq:lin_ref_ante}
		\bm v^a_{i,d}(\bm p_i) = \frac{\bm v_{i,\text{imp}} + \alpha(\bm p_{i,t}(\bm p_i) - \bm p_i)}{\lVert{\bm v_{i,\text{imp}} + \alpha(\bm p_{i,t}(\bm p_i) - \bm p_i)}\rVert_2}\lVert{\bm v_{i,\text{imp}}}\rVert_2,
	\end{equation}
	with user-defined shaping parameter $\alpha \in \mathbb{R}^+$ and intermediate target position $\bm p_{i,t}(\bm p_i)$ as
    \begin{equation}\label{eq:p_t}
		\bm p_{i,t}(\bm p_i) = \bm p_{i,\text{imp}} -  \frac{\bm v_{i,\text{imp}}\lVert{\bm p_{i,\text{imp}} - \bm p_i}\rVert_2}{\lVert{\bm v_{i,\text{imp}}}\rVert_2}.
	\end{equation}
    This will result in a vector field that can guide each end effector towards $\bm p_{i,\text{imp}}$ until the end effector moves past $\bm p_{i,\text{imp}}$, at which point ${\bm v}^a_{i,d}(\bm p_i)$ will guide the end effector back to $\bm p_{i,\text{imp}}$. 
    As we do not want the desired velocity to guide the end effector back to $\bm p_{i,\text{imp}}$ if uncertainties in the environment cause a delayed impact, an extension is required to ensure that there is always a suitable ante-impact reference that continues to guide the end effector towards the box. This extension is following the philosophy of classical time-based RS, where the reference is extended in time past the nominal impact time. The essential difference for time-invariant RS is that the proposed extension takes place in space past the nominal impact location (rather than in time beyond the nominal impact time). The extended velocity reference $\bar{\bm v}_{i,d}^a(\bm p_i)$ is determined via
	\begin{equation}\label{eq:dp_d}
	\bar{\bm v}_{i,d}^a(\bm p_i) = \beta^a_i(\bm p_i) \bm v^a_{i,d}(\bm p_i) + (1-\beta^a_i(\bm p_i))\bm v_{i,\text{imp}},
	\end{equation}
	with blending function $\beta^a_i(\bm p_i)$ defined as
	\begin{equation}
	\beta^a_i(\bm p_i) = S^a_1(\|\bm p_i - \bm{p}_{o,\text{est}}\|),
	\end{equation}
	where $\bm{p}_{o,\text{est}}$ is the estimated center of the box, and $S^a_1: \mathbb{R} \to \mathbb{R}$ is the first-order smoothstep function 
	\begin{equation} \label{eq:smoothstep}
	S^a_1(r) = 
	\left\{\begin{aligned}
	0 & \text { if } r \leq r^a_{\mathrm{min}}, \\
	3 r_w^2 - 2 r_w^3  & \text { if } r^a_{\mathrm{min}} <  r < r^a_{\mathrm{max}}, \\
	1 & \text { if } r \geq r^a_{\mathrm{max}} \\
	\end{aligned}\right.
	\end{equation}
	with user-defined $r^a_{\mathrm{min}}, r^a_{\mathrm{max}} \in \mathbb{R}^+$ such that $r^a_{\mathrm{max}} > r^a_{\mathrm{min}} > \|{\bm p_{i,\text{imp}} - \bm{p}_o}\|$ for $i \in \{1,2\}$, and
	\begin{equation}
	r_w = \frac{r - r^a_{\mathrm{min}}}{r^a_{\mathrm{max}} - r^a_{\mathrm{min}}}.
	%r_w = ({r - r^a_{\mathrm{min}}})({r^a_{\mathrm{max}} - r^a_{\mathrm{min}}})^{-1}.
	\end{equation}
	This creates two circles around the center of the object with radii $r^a_{\mathrm{min}}$ and $r^a_{\mathrm{max}}$, shown as dashed circles in Figure \ref{fig:ante_reference}. The area outside of the larger circle corresponds to the vector field ${\bm v}_{i,d}^a(\bm p_i)$ in \eqref{eq:lin_ref_ante} being followed, the area inside the smaller circle corresponds to a reference velocity $\bm v_{i,\text{imp}}$, and in between the two circles, a convex combination of $\bm v_{i,d}^a(\bm p_i)$ and $\bm v_{i,\text{imp}}$ is taken. The value for $r_{\mathrm{min}}^a$ is chosen as the largest possible distance between the center of the object and any of its corners in order to ensure that the desired velocity is equal to $\bm v_{i,\text{imp}}$ at all possible impact locations.
 
    A desired linear acceleration is then defined on the basis of $\bar{\bm v}^a_{i,d}(\bm p_i)$ as
    \begin{equation}\label{eq:a_a_d}
    \bar{\bm a}^a_{i,d}(\bm p_i) = \frac{\partial \bar{\bm v}^a_{i,d}}{\partial \bm p_i}(\bm p_i) \bar{\bm v}^a_{i,d}(\bm p_i),
    \end{equation}
    which corresponds to the nominal end effector acceleration if the reference of $\bar{\bm v}^a_{i,d}(\bm p_i)$ is followed perfectly. 
    An angular velocity reference, assuming a given constant desired orientation ${\bm R}_{i,d}$, is then defined as
    \begin{equation}\label{eq:omega_i_d}
    \bm \omega_{i,d}^a(\bm R_i) = -\kappa_r^a(\log(\bm R_i^T{\bm R}_{i,d}))^{\vee }
    \end{equation}
    with constant user-defined $\kappa_r^a$. An angular velocity reference $\bm \alpha_{i,d}^a(\bm R_i)$ is derived from \eqref{eq:omega_i_d} in similar fashion to \eqref{eq:a_a_d}. 
    The linear and angular velocity and acceleration vectors are then assembled into a twist and acceleration vector as 
    \begin{equation}\label{eq:vv_ante}
    \bar{\vv}^a_{i,d}(\bm p_i, \bm R_i)= [\bar{\bm v}^a_{i,d}(\bm p_i); {{\bm \omega}}^a_{i,d}(\bm R_i)]
    \end{equation}
    and
    \begin{equation}\label{eq:va_ante}
        \bar{\va}^a_{i,d}(\bm p_i, \bm R_i) = \left[ \bar{\bm a}^a_{i,d}(\bm p_i); \bm \alpha_{i,d}^a(\bm R_i) \right]. 
    \end{equation}
    A visualization of the ante-impact velocity field for a planar dual-arm grabbing scenario is depicted in Figure \ref{fig:ante_reference}. 
    Since the twist reference only prescribes 6DOF, an additional 1DOF reference is formulated to suppress any motion in the null space of the 7DOF robots used in this work, resulting in a predictable impact configuration. For this work, similar to \cite{Steen2024}, we select a constant desired position for the first joint of each robot $\xi_i := q_{i,1}$, defined as $\xi_{i,d}$.

\subsection{Post-impact velocity field}

    Since this work focuses on manipulation tasks with sustained contact after impact between robots and objects, we formulate a linear velocity reference for the object denoted by $\bar{\bm v}_{o,d}^p(\bm p_o)$ with $\bm p_o$ as the estimated position of the object\footnote{Because no vision component is included in our experimental validation, the position of the object $\bm p_o$ is inferred from the state of the robot(s), assuming the object and the robot are in contact.}. 
    Without loss of generality, we assume that the goal of the post-impact reference is to guide the object to a desired final pose denoted by $\bm p_{o,f}$ while maintaining a desired constant orientation ${\bm R}_{i,d}$ with the end effector(s). Still, as described in Section \ref{sec:introduction}, we aim to do so while ensuring that the ante- and post-impact references are compatible with the impact dynamics of the system in a neighborhood of the impact surface, minimizing velocity errors and input spikes after contact is established. To achieve this, the post-impact velocity is predicted through simulations performed using a nonsmooth physics engine. Section \ref{sec:impact_map} describes this prediction procedure, followed by the reference formulation in Section \ref{sec:post_ref_formulation}. %based on the state of the system when entering the post-impact mode and the ante-impact velocity reference.

\subsubsection{Impact-map formulation}\label{sec:impact_map}

\begin{figure}
		\centering
		\includegraphics[width=0.92\linewidth]{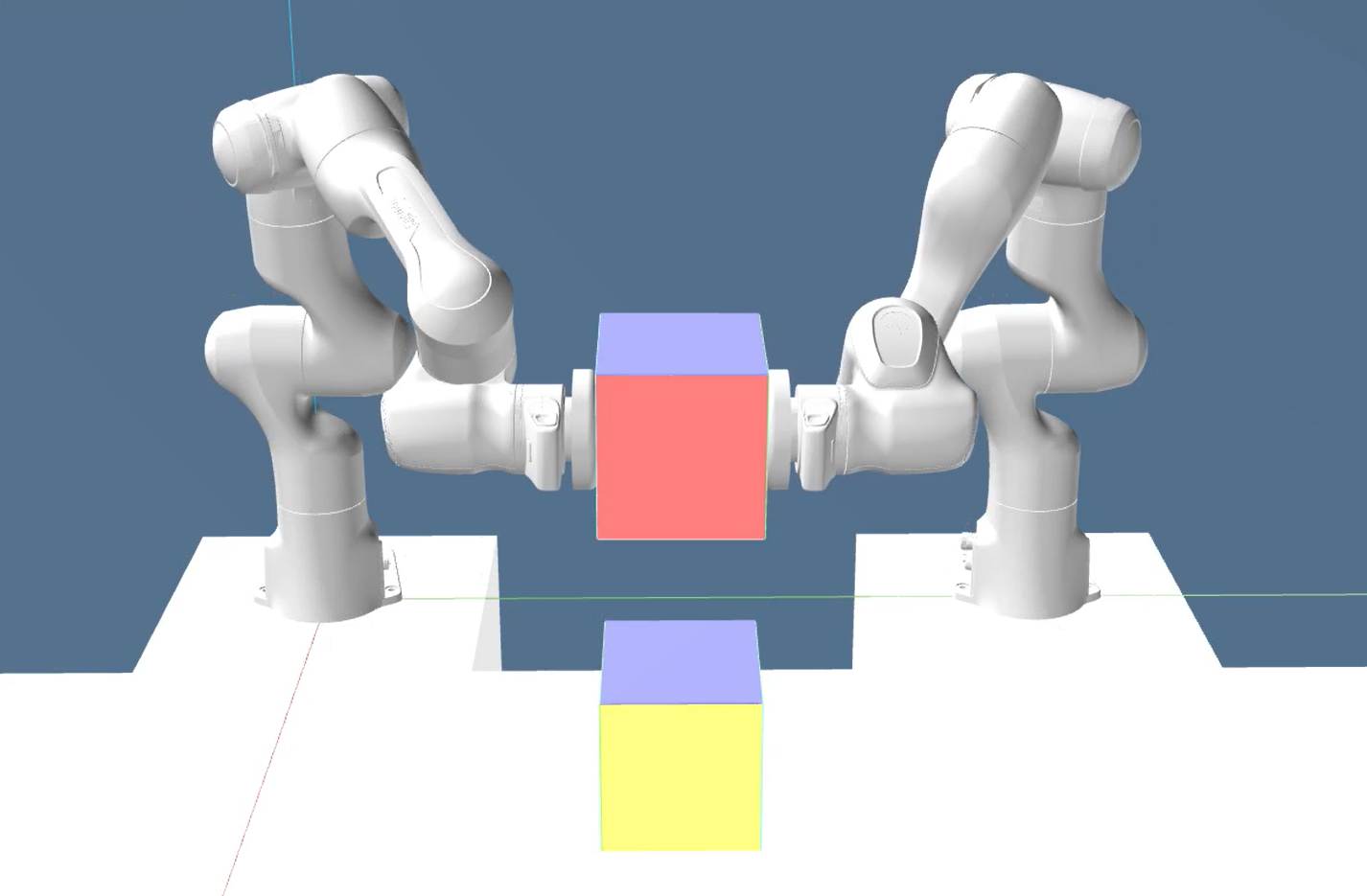}
		\caption{Visualization of the simulation environment for a dual-arm grabbing use case.}
		\label{fig:simulation}
	\end{figure}

    The simulations performed to determine the impact map for a range of possible impact states are performed using a nonsmooth physics engine. In particular, we employ AGX Dynamics\footnote{AGX Dynamics: \href{https://www.algoryx.se/agx-dynamics/}{https://www.algoryx.se/agx-dynamics/}}, which has recently seen increased interest in academia \cite{Wiberg2022,Li2020,Styrud2022,Cao2023} due to its fidelity and target of engineering applications. Following the assumption that sustained contact is achieved after the impact, the coefficient of restitution in these simulations is set to 0. Furthermore, following \eqref{eq:eom}, a motor inertia given by $\bm B_{\theta,i}$ is added to the simulations on top of the link inertia, described by $\bm M_i$. A screenshot of the virtual scene used for dual-arm grabbing simulations is given in Figure \ref{fig:simulation}, which replicates the existing physical setup shown in Figure \ref{fig:setup}. 
    
    In \cite{Steen2024a}, a procedure is described to determine the impact map for a given ante-impact robot configuration and velocity through the execution of a single simulation step, simulating the moment of impact. The impact map was then validated against real experiments for the hit-and-push and dual-arm grabbing motions also used in this work. However, a challenge in this work is that the nature of the ante-impact velocity field implies that, even if the velocity reference field is followed perfectly, the impact configuration, and with this the predicted post-impact velocity, varies depending on the initial configuration of the system. To overcome this challenge, simulations using the physics engine are performed for a range of possible impact locations %$\bm p_j^- = [\bm p_{1,j}^-; \bm p_{2,j}^-]$ 
    $\bm p_{i,j}^-$ and constant orientation ${\bm R}_{i,d}$ aligned with the object, with robot index $i$ and simulation index $j \in \{1, \dots, N_\text{exp}\}$, with $N_\text{exp}$ as the total amount of simulations performed\footnote{Because the robots used in this work contain 7DOF, we initialize the simulations with $q_{i,1} = \xi_{i,d}$
    and $\dot{q}_{i,1} = 0$, and infer the full ante-impact joint configuration and velocities through inverse kinematics.}. The ante-impact velocities $\bm v_{i,j}^-$ are given by the desired velocity field, i.e., $\bm v_{i,j}^- = \bar{\bm v}^a_{i,d}(\bm p_{i,j}^-)$, with angular velocities $\bm \omega_{i,j}^- = {\bm \omega}^a_{i,d}({\bm R}_{i,d}) = \bm 0$. 
    Simulations for all experiments are performed for a single 5 ms timestep under pure gravity compensation. The post-impact object velocity $\bm v_{o,j}^+$ is then extracted from the simulation for each simulated impact configuration $j$. 
    
    During physical experiments, the predicted impact state is extracted from the set of previously recorded simulations through a radial basis function (RBF) interpolation approach based on the impact configuration of the robot(s), resulting in a post-impact predicted post-impact velocity $\bm v^+_{o,\text{est}}$. This implies that the post-impact velocity reference field can be different between repetitions of the same experiment due to variations in the robot configuration at the time of impact. More details on the RBF interpolation procedure used to determine $\bm v^+_{o,\text{est}}$ can be found in Appendix A.

\subsubsection{Reference formulation}\label{sec:post_ref_formulation}

    %\paragraph{Linear velocity reference}
    
    With the goal of formulating a vector field that drives the object towards a final position $\bm p_{o,f}$, we now formulate a reference field for the velocity of the manipulated object $\bm v_o$. Because the reference should also be consistent with the ante-impact reference, %to avoid a large impact-induced error after full contact is established, 
    the desired linear velocity $\bar{\bm v}_{o,d}^p(\bm p_o)$ is obtained by merging two vector fields. The first vector field $\bm v^p_{o,d,f}$, which drives the object to a desired final position $\bm p_{o,f}$, is given by
    \begin{equation} \label{eq:attractor}
        \bm v^p_{o,d,f}(\bm p_o) = \kappa^p_p (\bm p_{o,f} - \bm p_{o}),
    \end{equation}
    with user-defined $\kappa^p_p \in \mathbb{R}^+$. The second vector field is given by the nominal post-impact box velocity $\bm v^+_{o,\text{est}}$. The two vector fields are blended to obtain the extended velocity reference field
    \begin{equation} \label{eq:post_ext_vel}
    \bar{\bm v}_{o,d}^{p}(\bm p_{o}) = \beta^{p}(\bm p_{o})\bm v_{o,d,f}^{p}(\bm p_{o}) + (1-\beta^{p}(\bm p_{o}))\bm v^+_{o,\text{est}}.
    \end{equation}
    The blending function $\beta^{p}(\bm p_{o})$ is given by
    \begin{equation} \label{eq:post_transition}
        \beta^{p}(\bm p_{o}) = S^p_1\left(||{\bm p^+_{o} - \bm p_o}||\right)
    \end{equation}
    with smoothstep $S^p_1(r)$ defined as in \eqref{eq:smoothstep} with $r^p_\text{min}  < r^p_\text{max} \leq ||\bm p^+_{o} - \bm p_{o,f}||$,
    and $\bm p^+_{o}$ as the object position at the time of impact. By its definition, the post-impact reference is equal to the predicted post-impact velocity $\bm v^+_{o,\text{est}}$ at the moment where the impact is detected, thus resulting in a minimal tracking error and minimal resulting control action. A visualization of a possible desired post-impact velocity field for a 2D dual-arm grabbing scenario is presented in Figure \ref{fig:post_field}, also showing the contribution of the two components $\beta^{p}(\bm p_{o})\bm v_{o,d,f}^{p}(\bm p_{o})$ and $(1-\beta^{p}(\bm p_{o}))\bm v^+_{o,\text{est}}$. The path taken if the reference $\bar{\bm v}_{o,d}^{p}(\bm p_{o})$ is followed perfectly is shown in Figure \ref{fig:post_path} together with the path taken if $\bm v_{o,d,f}^{p}(\bm p_{o})$ would be followed. The more natural path that is followed with the reference field $\bar{\bm v}_{o,d}^{p}(\bm p_{o})$ shows the positive effect of including $v^+_{o,\text{est}}$ in the reference formulation. 
    
    The desired angular velocity of the robots is formulated as
    \begin{equation}
    \bm \omega_{i,d}^p(\bm R_i) = \kappa_r^p(\log(\bm R_i^T{\bm R}_{i,d}))^{\vee}
    \end{equation}
    with constant desired impact orientation ${\bm R}^p_{i,d}$. 
    %\paragraph{Angular velocity reference}
    Since we aim to maintain a constant orientation and since we assume sustained non-slipping contact after the impact, it follows without loss of generality that the velocity reference for the robot(s) is identical to the reference of the box, i.e., $\bar{\bm v}^p_{i,d}(\bm p_o) = \bar{\bm v}^p_{o,d}(\bm p_o)$.  
    The linear and angular velocity are then collected into a single twist vector $\bar{\vv}^p_{i,d}(\bm p_o, \bm R_i)$ as in \eqref{eq:vv_ante}, and an acceleration vector is defined from $\bar{\vv}^p_{i,d}(\bm p_o, \bm R_i)$ to define $\bar{\vv}^p_{i,d}(\bm p_o, \bm R_i)$ in similar fashion as \eqref{eq:va_ante}.
    
    %A desired acceleration $\bar{\bm a}^p_{o,d}$ is then defined from $\bar{\bm v}^p_{o,d}(\bm p_o)$ in similar fashion to \eqref{eq:a_a_d}.

\section{Controller design}
\label{sec:control}

    Using the velocity reference fields defined in Section \ref{sec:reference}, a controller is constructed for an ante-impact, an interim, and a post-impact mode according to the time-invariant reference spreading framework. For each of the three modes, a QP controller is designed. 
	The switching policy between the modes is based on detection of the first impact of either robot, which activates the interim mode that remains active for a set amount of time $\Delta t_\text{int}$, followed by the start of the post-impact mode. The impact detection procedure used in this work is identical to the approach desribed in \cite{Steen2024}. Each of the three modes has a corresponding QP controller with optimization variables $\ddot{\bm q}_i$, which is solved at fixed time intervals $\Delta t$ to compute an optimal desired joint acceleration $\ddot{\bm q}^*_i$. This acceleration is converted into a desired joint torque $\bm \tau^*_i$ by using the equations of motion of the robot(s) \eqref{eq:eom} in free motion, i.e., 
	\begin{equation}\label{eq:EOM_free}
	\bm \tau^*_i = \left(\bm M_i + \bm B_{\theta,i}\right)\ddot{\bm q}^*_i + \bm h_i.
	\end{equation}
    In the following, the QP formulation in each of the control modes will be discussed.

\subsection{Ante-impact controller}

    In the ante-impact mode, a task on acceleration level is used to have the end effector track the desired twist $\bar{\vv}^a_{i,d}(\bm p_i,\bm R_i)$ from \eqref{eq:vv_ante}. The corresponding error for this task is defined as 
    \begin{equation} \label{eq:e_imp_ante}
    	\bm e^a_{i,\text{track}} :=  \va_i - \va^a_{i,t}
    \end{equation}
    %with task-space equivalent inertia matrix $\bm \Lambda_i$ as
    %\begin{equation}
    %	\bm \Lambda_i := \left(\bm J_i\bm M_i^{-1} \bm J_{i}^T\right)^{-1}       
    %\end{equation}
    with target acceleration $\va^a_{i,t}$ as
    \begin{equation}\label{eq:a_a_t}
    \va^a_{i,t} = \bar{\va}^a_{i,d}(\bm p_i,\bm R_i) + \bm D_\text{track} \left( \bar{\vv}^a_{i,d}(\bm p_i,\bm R_i) - \vv_i \right)
    \end{equation}
    and diagonal user-defined gain matrix $\bm D_\text{track} \in \mathbb{R}^{6 \times 6}$. 
    Bringing the task error from \eqref{eq:e_imp_ante} to 0 implies that $\va_i = \va^a_{i,t}$, which results in a velocity feedback effort to follow the desired velocity field with added acceleration feedforward. In a similar fashion, a task tracking the desired position of the first joint $\xi_{i}$ is defined to resolve kinematic redundancy as highlighted in Section \ref{sec:ante_ref_formulation}, with error $e^a_{i,q}$ as
    \begin{equation}\label{eq:e_q_ante}
         e^a_{i,q} :=  \ddot{\xi}_i + 2 \sqrt{k_q}\dot{\xi}_i - k_q({\xi}_{i,d} - \xi)
    \end{equation}
    and user-defined gain $k_q \in \mathbb{R}$. 
    
    While these two tasks suffice to perform a single-arm impacting motion such as the hit-and-push use case, an additional task is required for dual-arm manipulation if a simultaneous impact is planned. Simply following the reference $\bar{\va}^a_{i,d}(\bm p_i,\bm R_i)$ for both robots might result in a large mismatch in impact timing between the left and right arm.    
    To encourage synchronization between the end effectors with respect to the object, we propose to mirror the end effector positions across a plane that is parallel with and equidistant to the impact surfaces of the object, which are assumed to be parallel to each other. This way, we can compare $\bm p_1$ with the mirrored version of $\bm p_2$ and vice versa. These mirrored positions $\bm p_{1,m}$ and $\bm p_{2,m}$ are given by
    \begin{equation} \label{eq:sync_point1}
    \bm p_{1,m} = \bm T_{s}(\bm p_{1} - \bm p_{o}) + \bm p_{o}
    \end{equation}
    \begin{equation} \label{eq:sync_point2}
    \bm p_{2,m} = \bm T_{s}(\bm p_{2} - \bm p_{o}) + \bm p_{o}
    \end{equation}
    with estimated center of the object $\bm p_{o}$ and
    \begin{equation}\label{eq:T_s}
    \bm T_s = \bm I - 2 \bm n_1 \bm n_1^T
    \end{equation}
    with $\bm n_i$ as the vector normal to the contact surface of the object with which robot $i$ is to impact. Under the assumption that both contact surfaces of the object are parallel, either $\bm n_1$ or $\bm n_2$ can be used in \eqref{eq:T_s}. Assuming the object is stationary in the ante-impact mode, the mirrored velocity and acceleration are given by
	\begin{equation*} \label{eq:sync_vel}
	\bm v_{1,m} = \bm T_{s}\bm v_{2}, \ \ \ \bm v_{2,m} = \bm T_{s}\bm v_{1}. %\ \ \ \ddot{\bm p}_{2,m} = \bm T_{s}(\bm J_{p,2}\ddot{\bm q}_2 + \dot{\bm J}_{p,2}\dot{\bm q}_2).
	\end{equation*}
	Using this definition, the errors $\bm e^a_{1,\text{sync}}$ and $\bm e^a_{2,\text{sync}}$ can be defined as
	\begin{equation}
	\bm e^a_{1,\text{sync}} := \ddot{\bm p}_{1} - k_\text{sync}({\bm p}_{2,m}-{\bm p}_{1}) - 2\sqrt{k_\text{sync}}(\bm v_{2,m}-\bm v_{1}),
	\end{equation}
    \begin{equation}
	\bm e^a_{2,\text{sync}} := \ddot{\bm p}_{2} - k_\text{sync}({\bm p}_{1,m}-{\bm p}_{2}) - 2\sqrt{k_\text{sync}}(\bm v_{1,m}-\bm v_{2})
	\end{equation}
	with user-defined gain $k_\text{sync} \in \mathbb{R}^+$. %Enforcing $\bm e^a_{s}=\bm 0$ results in synchronization between both robots. 
    The QP control cost function for the dual-arm grabbing case is then determined by taking a weighted sum of these errors as
    \begin{equation}\label{eq:E_ante}
        \begin{aligned}
        E_\text{ante} = \sum_{i=1}^2 \big(&w_\text{track}\|\bm e^a_{i,\text{track}}\|^2 + w_\text{q}e_{i,q}^2 + \\ &w_\text{sync} \|{\bm p}_{2,m}-{\bm p}_{1}\|\|{\bm e^a_{i,\text{sync}}}\|^2\big), 
        \end{aligned}
    \end{equation}
	with user-defined weights $w_\text{track}, w_q ,w_\text{sync} \in \mathbb{R}^+$. Note that the synchronization error is scaled with $\|{\bm p}_{2,m}-{\bm p}_{1}\|$ to ensure the effect of the synchronization task vanishes as the manipulators converge to a mirrored position. 
	%\begin{equation}
	%\begin{aligned}
	%E_\text{ante} & =  w_p\sum_{i=1}^2\left(\ddot{\bm q}_i^T \bm J_{p,i}^T \bm J_{p,i} \ddot{\bm q}_i + 2{\bm \eta^a_{p,i}}^T \bm J_{p,i} \ddot{\bm q}_i\right) \\ & + w_\theta\sum_{i=1}^2\left( \ddot{\bm q}_i^T \bm J_{\theta,i}^T \bm J_{\theta,i}\ddot{\bm q}_i + 2\eta^a_{\theta,i} \bm J_{\theta,i} \ddot{\bm q}_i\right) \\ & + w_s \left( \right)
	%\end{aligned}
	%\end{equation}
    We combine this cost function with a set of constraints to prevent violations of the upper and lower limits of the joint positions, velocities and torques, defined as ${\bm q}_\text{max}, {\dot{\bm q}}_\text{max}$, ${\bm \tau}_\text{max}$ and ${\bm q}_\text{min}, {\dot{\bm q}}_\text{min}$, ${\bm \tau}_\text{min}$, respectively. This results the full ante-impact QP
	\begin{equation}\label{eq:QP_ante}
	(\ddot{\bm q}_1^*, \ddot{\bm q}_2^*) = \underset{\ddot{\bm q}_1, \ddot{\bm q}_2}{\operatorname{argmin}} \ E_\text{ante},
	\end{equation}
	s.t., for $i\in\{1,2\}$,
        \begin{equation}\label{eq:const_q}
	{\bm q}_\text{min} \leq \frac{1}{2}\ddot{\bm q}_i \Delta t^2 + \dot{\bm q}_i\Delta t + \bm q_i \leq {\bm q}_\text{max},
	\end{equation} 
	\begin{equation}\label{eq:const_dq}
	\dot{\bm q}_\text{min} \leq \ddot{\bm q}_i \Delta t + \dot{\bm q}_i \leq \dot{\bm q}_\text{max},
	\end{equation} 
	\begin{equation}\label{eq:const_tau}
	{\bm \tau}_\text{min} \leq (\bm M_i+\bm B_{\theta,i})\ddot{\bm q}_i + \bm h_i  \leq {\bm \tau}_\text{max}
	\end{equation} 
	with QP time step $\Delta t$. 
    The reference torques $\bm \tau_1^*$ and $\bm \tau_2^*$ are then obtained from $\ddot{\bm q}_1^*$ and $\ddot{\bm q}_2^*$ using \eqref{eq:EOM_free}. Note that the QP defined for the hit-and-push use case is nearly identical to this dual-arm grabbing QP, with the only differences being that all tasks and constraints are defined for a single robot, and the synchronization task (last term in \eqref{eq:E_ante}) is removed.

    \begin{figure*}
		\centering
		\begin{subfigure}[b]{0.241\textwidth}
			\centering
			\includegraphics[width=\textwidth]{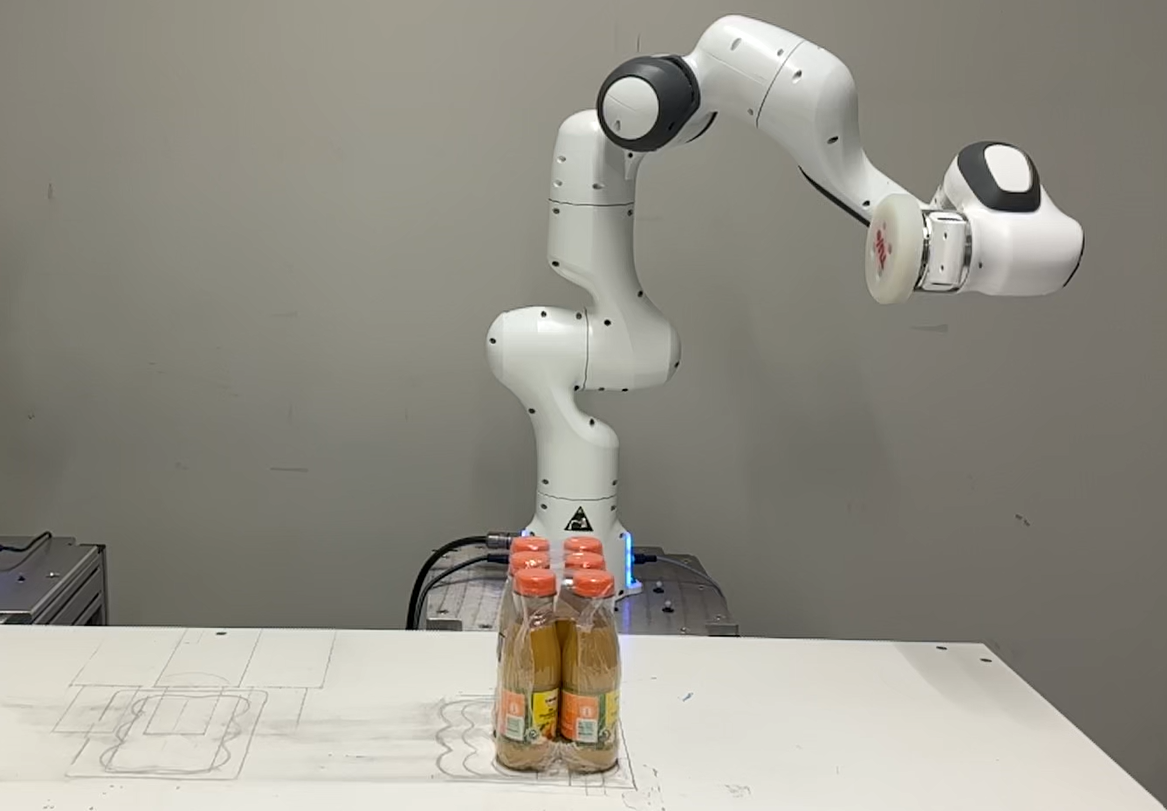}
			\caption{Initial configuration}
			\label{fig:Push_sequence_1}
		\end{subfigure}
		\hfill
		\begin{subfigure}[b]{0.24\textwidth}
			\centering
			\includegraphics[width=\textwidth]{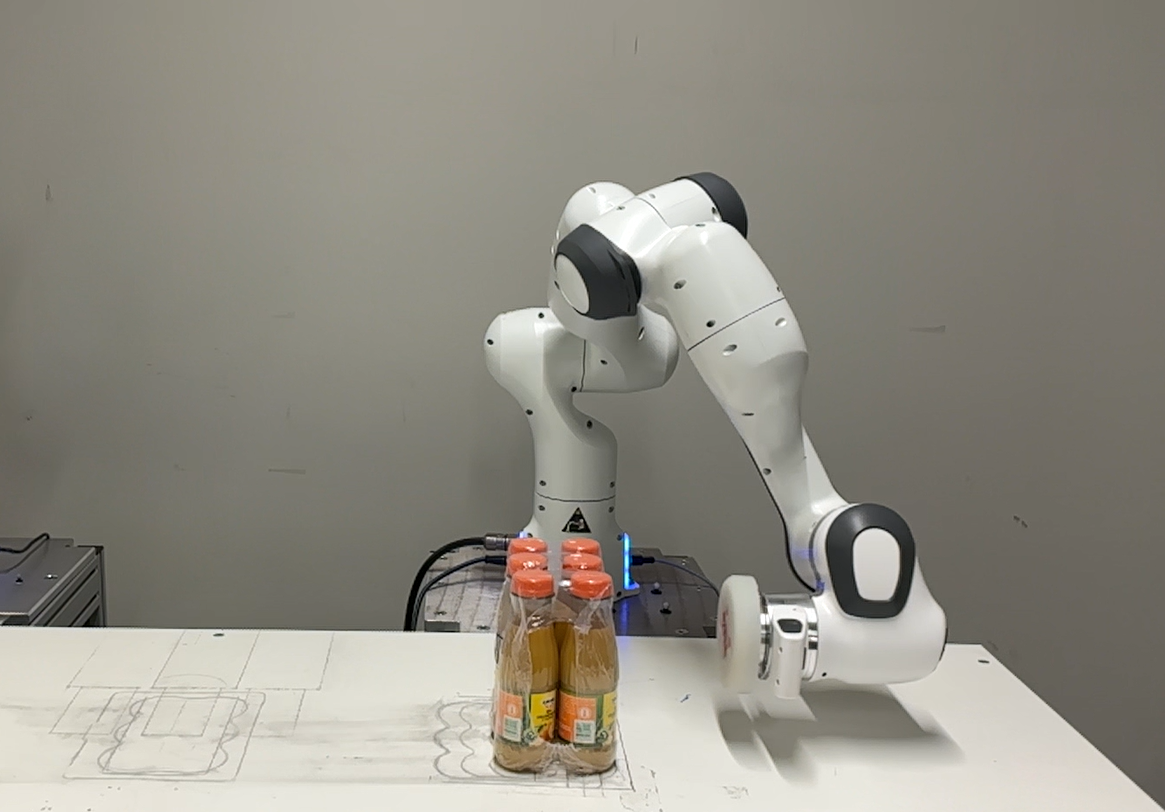}
			\caption{Align with impact direction}
			\label{fig:Push_sequence_2}
		\end{subfigure}
		\hfill
		\begin{subfigure}[b]{0.2415\textwidth}
			\centering
			\includegraphics[width=\textwidth]{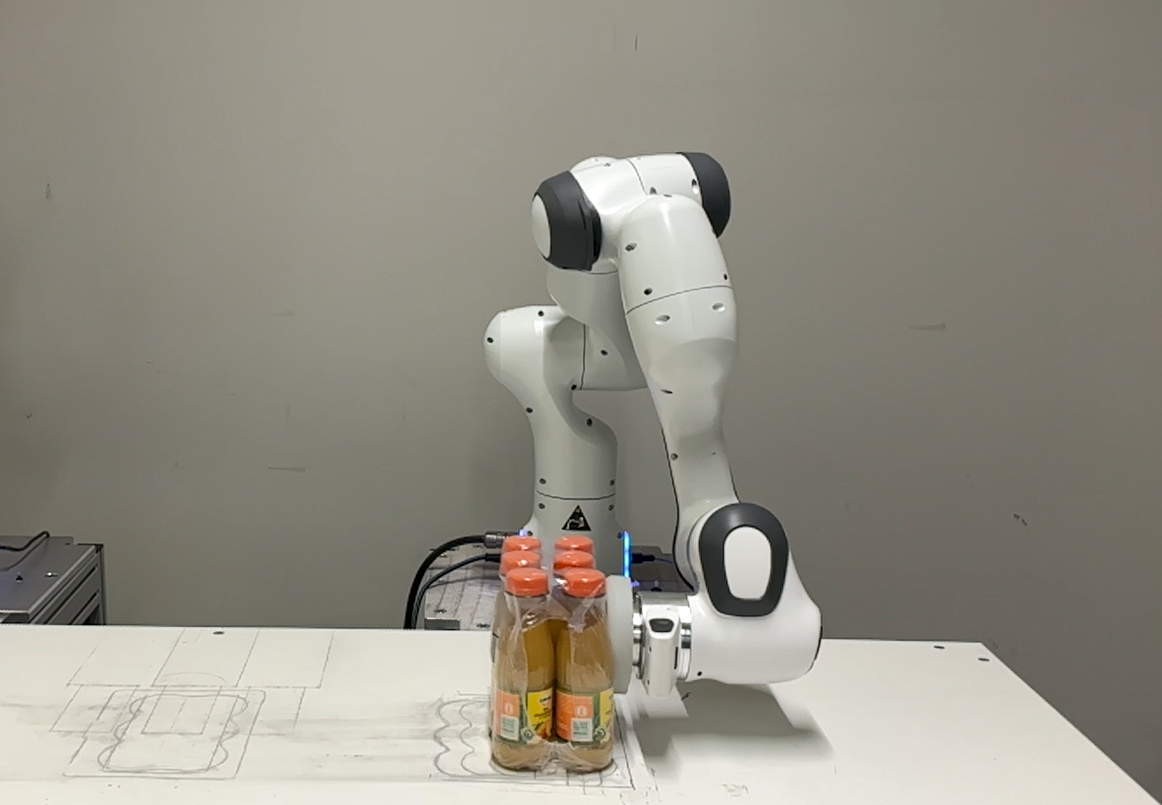}
			\caption{Impact configuration}
			\label{fig:Push_sequence_3}
		\end{subfigure}
        \hfill
		\begin{subfigure}[b]{0.242\textwidth}
			\centering
			\includegraphics[width=\textwidth]{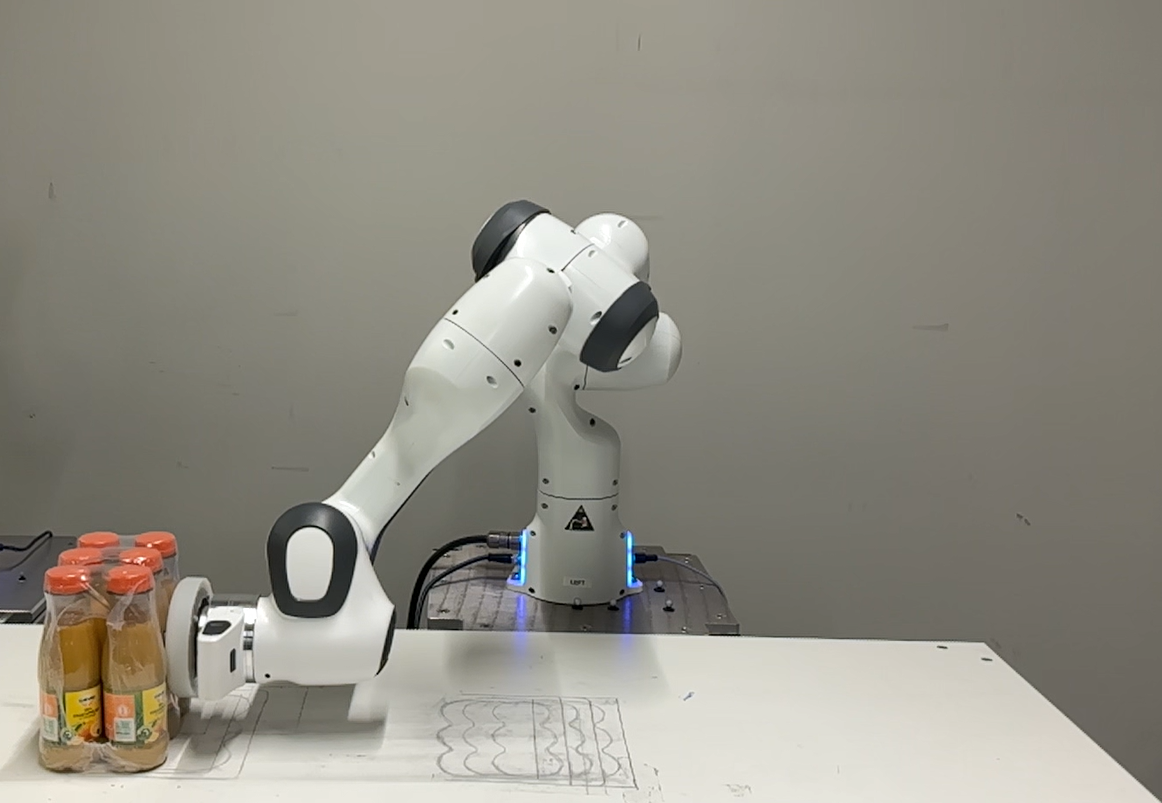}
			\caption{Post-impact configuration}
			\label{fig:Push_sequence_4}
		\end{subfigure}
    \caption{Snapshots of the system for one of the hit-and-push experiments. In (a), the selected initial configuration is shown. In (b), it is shown that the ante-impact velocity field is followed to align the end effector with the impact direction before impacting the object as shown in (c). After following a velocity field that is adjusted to the predicted impact map, (d) shows the final configuration of the system after the object is pushed to its desired location.}
	\label{fig:snapshots_push}
	\end{figure*}

\subsection{Interim controller}

    As highlighted in Section \ref{sec:introduction}, uncertainty during events with nominally simultaneous impacts inevitably leads to a small time separation between these impacts, in which the system resides neither in the ante-impact, nor the post-impact velocity state. The interim mode QP controller, which is activated upon detection of the first impact, is tasked with ensuring contact completion without inducing input peaks. This is achieved by initially removing velocity feedback and replacing it by a position feedback effort based on the ante-impact velocity reference field. %, in order to drive the end effector to establish sustained contact without peaks induced by the velocity error. The interim mode starts at the moment the first impact is detected and lasts for a pre-defined duration $\Delta t_\text{int}$.
    To derive this effort, a desired position $\bm p_{i,d}^\text{int}$ is derived from the ante-impact velocity reference $\bar{\bm v}^a_{i,d}(\bm p_i)$ through numerical integration as
    \begin{equation}\label{eq:pos_int}
	\bm p_{i,d}^\text{int}(t_k + \Delta t) = \bm p_{i,d}^\text{int}(t_k) + \bar{\bm v}^a_{i,d}(\bm p_{i,d}^\text{int}(t_k)) \Delta t
	%\theta_d^\text{int}(t_k + \Delta t) &= \theta_d^\text{int}(t_k) + \dot{\theta}^a_d(\theta_d^\text{int}(t_k)) \Delta t,
    \end{equation} 
    %\begin{equation}\label{eq:ori_int}
	%\bm R_{i,d}^\text{int}(t_k + \Delta t) = \bm R_{i,d}^\text{int}(t_k) \ \text{exp}\left(\bm \omega^a_{i,d}\left(\bm R_{i,d}^\text{int}(t_k)\right)\Delta t \right)
    %\end{equation}
    with QP time step $\Delta t$ and $\bm p_{i,d}^\text{int}(T_\text{imp}) = \bm p_i(T_\text{imp})$, where $T_\text{imp}$ is the time where the interim mode is entered. %and $\bm R_{i,d}^\text{int}(t_\text{int}) = \bm R_i(t_\text{int})$. 
    Using these quantities, the tracking error in the interim mode is given by
    \begin{equation} \label{eq:e_int}
    	\bm e^\text{int}_{i,\text{track}} :=  \va_i - \va^\text{int}_{i,t}.
    \end{equation}
    The target acceleration $\va^\text{int}_{i,t}$ is given by
    \begin{equation}\label{eq:a_int_t}
    \resizebox{\linewidth}{!}{$
	\begin{aligned}
	& \va^\text{int}_{i,t}  :=  (1-\gamma(t))\left( \bar{\va}^a_{i,d}(\bm p_i,\bm R_i) + \bm K_\text{track} \begin{bmatrix} {\bm p}^\text{int}_{i,d}(t) - {\bm p_{i}} \\ (\log(\bm R_i^T \bm R_{i,d}))^{\vee } \end{bmatrix} \right)
    \\ & +   \gamma(t)\left(\bar{\va}^p_{i,d}(\bm p_o,\bm R_i) + \bm \Lambda_i^{-1}\vf^p_{i,t}  +  \bm  D_\text{track} (  \bar{\vv}^p_{i,d}(\bm p_o,\bm R_i) - \vv_i )\right)
	\end{aligned}
    $}
    \end{equation}
    with task-space inertia matrix $\bm \Lambda_i$ and feedforward post-impact wrench $\vf^p_{i,t}$ both highlighted further in Section \ref{sec:control_post}, as well as user-defined gain $\bm K_\text{track}$ and so-called blending function $\gamma(t)$ defined as
    \begin{equation}
	\gamma(t) := \left(t-T_\text{imp}\right)/\Delta t_\text{int}.
    \end{equation}
    with $\Delta t_\text{int}$ as the user-defined duration of the interim mode (typically in the order of 100-200 ms). This blending procedure ensures a smooth transition without input jumps between the ante-impact mode and the interim mode, as well as between the interim mode and the post-impact mode. When $\gamma(t) = 0$ at the start of the interim mode, all velocity feedback is removed, reducing input peaks that would otherwise result from the inevitable rapid velocity changes induced by impacts. This leaves only the ante-impact feedforward acceleration term and a pose feedback term, split into a position and orientation feedback. The position feedback term is 0 by the definition of $\bm p_{i,d}^\text{int}$ in \eqref{eq:pos_int} at the start of this interim mode, but as time progresses, $\bm p_{i,d}^\text{int}$ will move into the impacted object, with the position feedback signal promoting contact completion as a result. %The orientation feedback term is also 0 at the start of the interim mode, assuming the end effector orientation has successfully converged to $\bm R_{i,d}$ in the ante-impact mode. 
    Over time, $\gamma(t)$ will gradually increase to $1$, at which time the task error in the QP controller is equivalent to the task error in the post-impact mode, as will become apparent in Section \ref{sec:control_post}. In doing so, an undesired jump in the input signals at this switching time is prevented.  
    Using $\gamma(t)$, we also redefine the task on the position of the first joint from \eqref{eq:e_q_ante} as
	\begin{equation} \label{eq:beta_pos_int}  
    e^\text{int}_{i,q} = \ddot{\xi}_i + 2\gamma(t)\sqrt{k_q} \dot{\xi}_i - k_q(\bar{\xi}_{i,d} - \xi),
	\end{equation}
    initially removing and gradually increasing the velocity feedback to reduce the effect of the velocity jump towards peaks in the control effort. 
	The full interim mode QP controller is then formulated similar to \eqref{eq:QP_ante} using the same constraints, and an objective function formed using $\bm e^\text{int}_{i,\text{track}}$ and $e^\text{int}_{i,q}$, without an added synchronization task.

    \begin{figure*}
		\centering
		\begin{subfigure}[b]{0.32\textwidth}
			\centering
			\includegraphics[width=\textwidth]{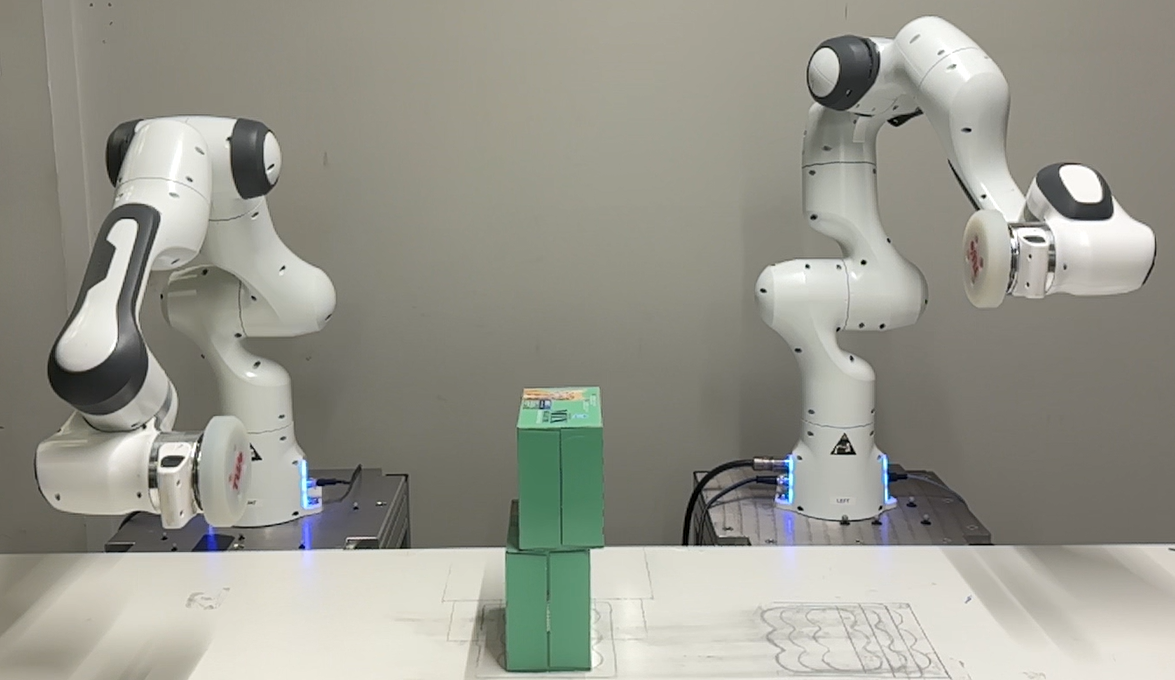}
			\caption{Initial configuration}
			\label{fig:Grab_sequence_1}
		\end{subfigure}
		\hfill
		\begin{subfigure}[b]{0.32\textwidth}
			\centering
			\includegraphics[width=\textwidth]{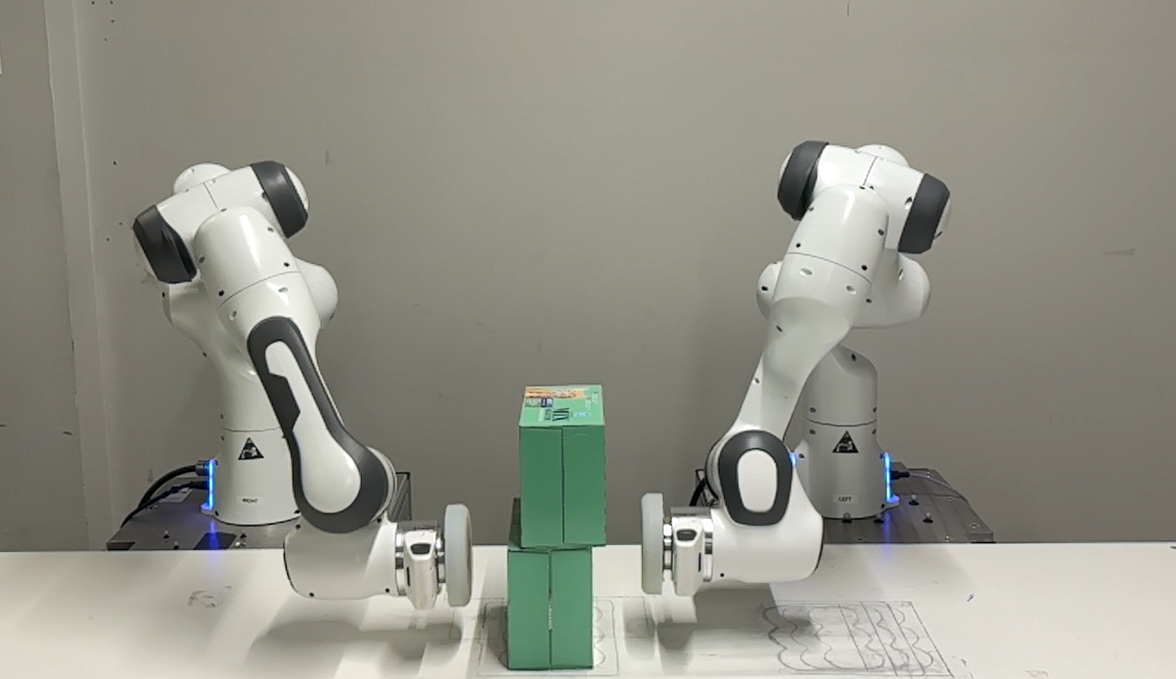}
			\caption{Sync and align with impact direction}
			\label{fig:Grab_sequence_2}
		\end{subfigure}
		\hfill
		\begin{subfigure}[b]{0.32\textwidth}
			\centering
			\includegraphics[width=\textwidth]{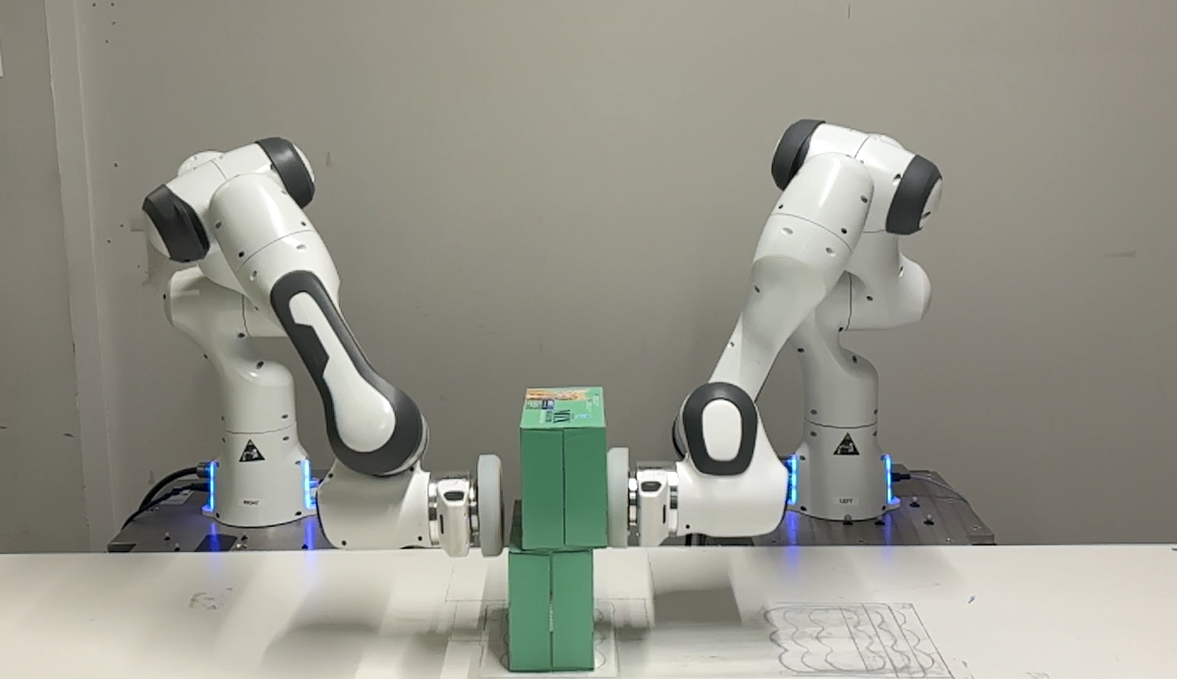}
			\caption{First impact \& start interim mode}
			\label{fig:Grab_sequence_3}
		\end{subfigure}
        \hfill
		\begin{subfigure}[b]{0.32\textwidth}
			\centering
			\includegraphics[width=\textwidth]{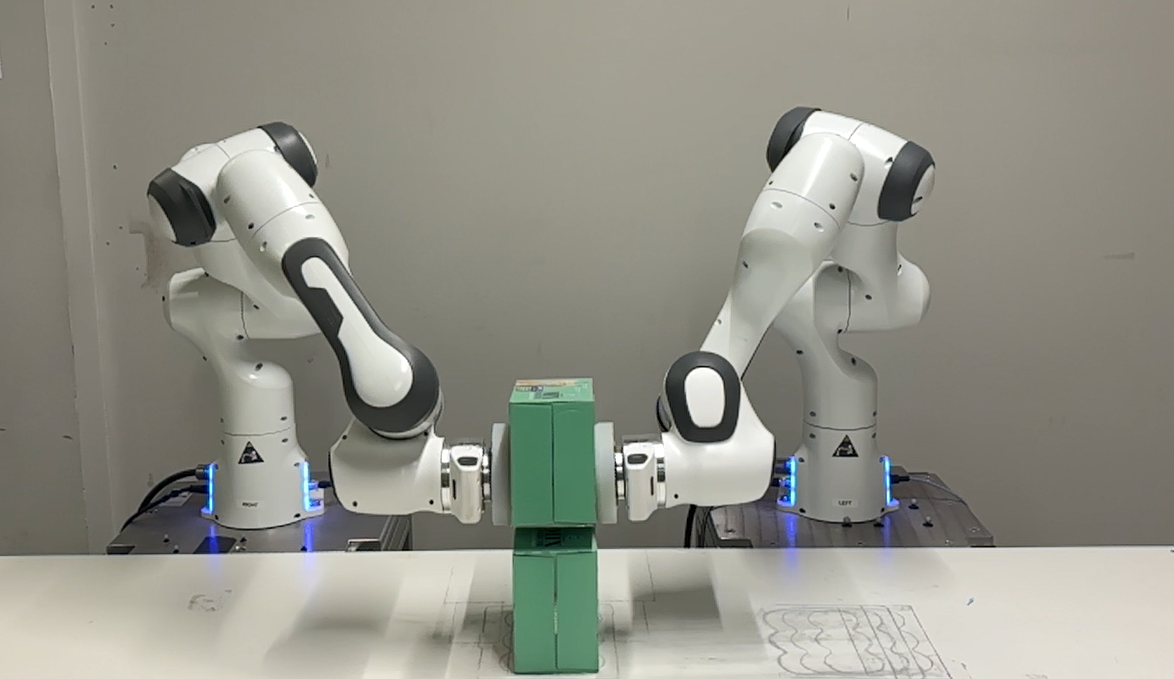}
			\caption{End interim mode: impact completed}
			\label{fig:Grab_sequence_4}
		\end{subfigure}
        \hfill
		\begin{subfigure}[b]{0.32\textwidth}
			\centering
			\includegraphics[width=\textwidth]{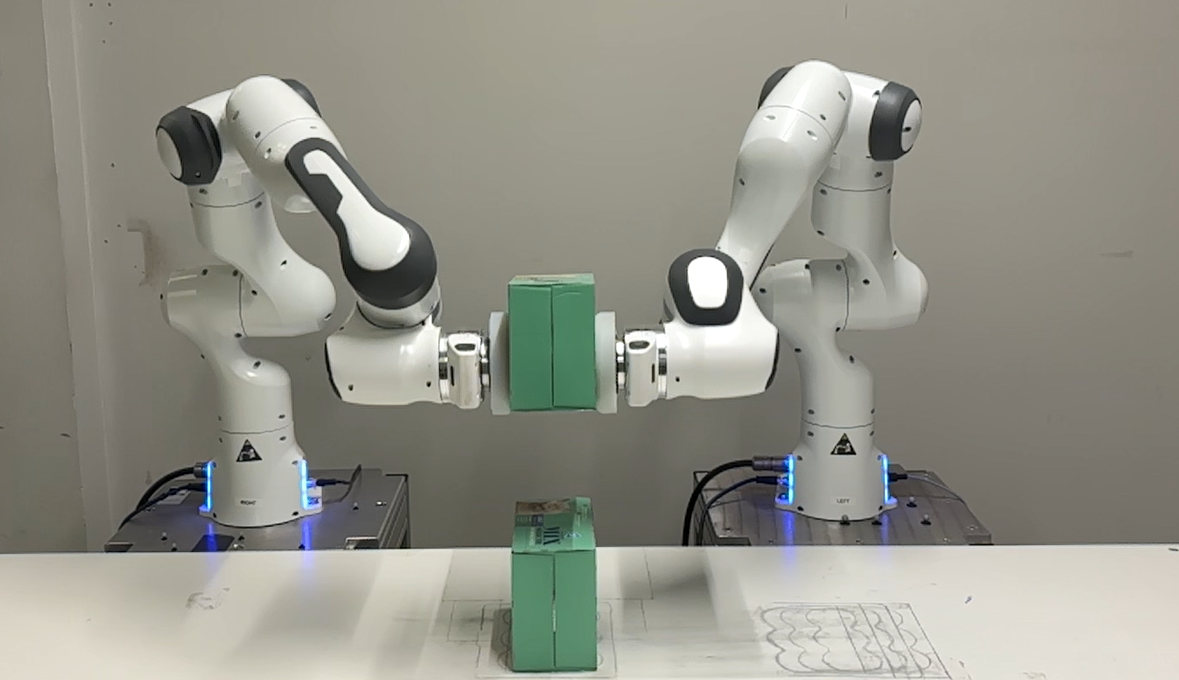}
			\caption{Path adjusted to predicted impact map}
			\label{fig:Grab_sequence_5}
		\end{subfigure}        \hfill
		\begin{subfigure}[b]{0.32\textwidth}
			\centering
			\includegraphics[width=\textwidth]{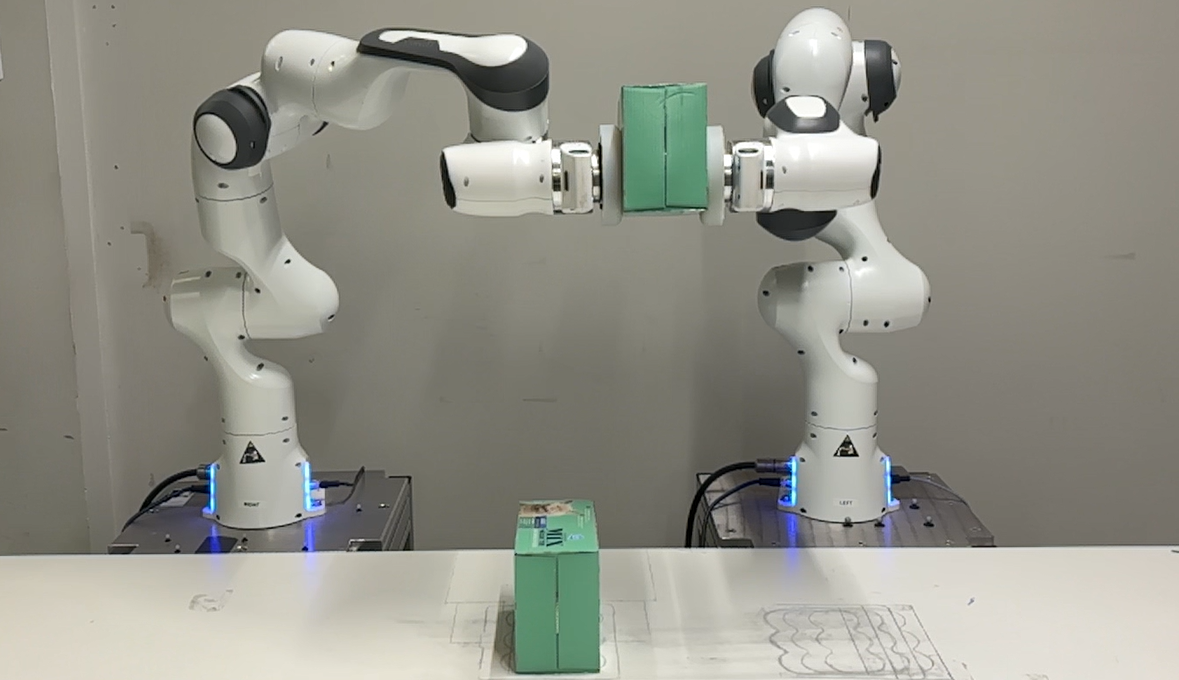}
			\caption{Final desired post-impact state}
			\label{fig:Grab_sequence_6}
		\end{subfigure}
        \caption{Snapshots of the system for one of the dual-arm grabbing experiments. In (a), the initial configuration is shown, with no symmetry between the two robots with respect to the object. In (b), it is shown that the robots synchronize while following the ante-impact velocity field to create a simultaneous impact between the two robots and the object. Due to an unmodeled 15mm displacement of the object, the right robot impacts the object first in (c), which triggers the interim mode. In (d), it is shown that the impact sequence is eventually successfully resolved, with (e) then showing that the path followed by tracking the post-impact velocity reference matches the direction of the predicted post-impact velocity. In (f), it is shown that the same post-impact velocity reference eventually steers the object to a desired post-impact rest pose.}
		\label{fig:snapshots_grab}
	\end{figure*}

\subsection{Post-impact controller}\label{sec:control_post}

    In the post-impact mode, an end effector task on acceleration level is defined that is similar to the ante-impact mode, with an additional contribution from the desired wrench applied at the end effector. The updated acceleration task is given by 
    \begin{equation} \label{eq:e_imp_post}
    	\bm e^p_{i,\text{imp}} :=  \va_i - \va^p_{i,t},
    \end{equation}
    with target acceleration $\va^p_{i,t}$ as
    \begin{equation}\label{eq:a_p_t}
    \va^p_{i,t}= \bar{\va}^p_{i,d}(\bm p_o,\bm R_i) + \bm \Lambda_i^{-1}\vf^p_{i,t} + \bm D_\text{track} \left( \bar{\vv}^p_{i,d}(\bm p_o, \bm R_i) - \vv_i \right),
     \end{equation}
     %with $\bm p_b$ estimated as
     %\begin{equation}
     %    \bm p_b = \frac{1}{2}(\bm p_1 + \bm p_2).
     %\end{equation}
    following the same structure as \eqref{eq:e_imp_ante} with an additional feedforward target wrench $\vf^p_{i,t}$ that is projected into an acceleration command through the equivalent task-space inertia matrix $\bm \Lambda_i$, given by 
    \begin{equation}
	\bm \Lambda_i := \left(\bm J_i(\bm M_i + \bm B_{\theta,i})^{-1} \bm J_{i}^T\right)^{-1}.    
	\end{equation}
    The desired wrench is tasked to compensate for the interaction forces between the end effector and the manipulated object. In the hit-and-push scenario, $\vf^p_{i,t}$ compensates the estimated friction force acting on the object due to the sliding, assuming Coulomb friction with a known friction coefficient. In the dual-arm grabbing case, $\vf^p_{i,t}$ contains the desired clamping force and compensation for the gravity acting on the lifted object. This task, together with a task to resolve redundancy identical to \eqref{eq:e_q_ante}, is used to create the post-impact QP with the same constraints as in the ante-impact and interim modes.
    %\begin{equation}
    %    \vf^p_{i,t} = [\bm f^p_{i,t}; 0; 0; 0]
    %\end{equation}
    %with 
    %\begin{equation}
    %    \bm f^p_{i,t} = \bm n_i f_{c} + \frac{1}{2}\bm M_b\left(\bm a_{b,d} + \bm g\right)
    %\end{equation}
    %with $f_c$ as the desired contact force, $\bm n_i$ as the contact force normal of robot $i$ with the box, and $\bm a_{b,d}$ as the desired box acceleration. 

    \section{Experimental validation}
    \label{sec:validation}

    As highlighted in Section \ref{sec:introduction}, this work shows for the first time an experimental validation of the time-invariant reference spreading control framework, which was previously validated only through simulations for a planar use case \cite{Steen2022a,Steen2022b}. We performed a total of 600 experiments in this experimental validation, and implemented the control approach described in Section \ref{sec:control} in the mc\_rtc\footnote{mc\_rtc: \url{https://jrl.cnrs.fr/mc_rtc/}} QP control software framework. The experiments are divided into a hit-and-push use case, shown in Figure \ref{fig:snapshots_push}, and a dual-arm grabbing use case, shown in Figure \ref{fig:snapshots_grab}.     
    Experiments for both use cases are performed using three objects with different sizes, shapes and materials, namely 1) a 0.60 kg foam-filled parcel (shown in Figure \ref{fig:setup}); 2) a 1.30 kg box of cat food (shown in Figure \ref{fig:snapshots_grab}); 3) a 2.20 kg pack of 6 juice bottles (shown in Figure \ref{fig:snapshots_push}). The parameter values used for all of the experiments are highlighted in Table \ref{tab:parameters}.

    \begin{table}[]
    \caption{Table containing parameter values used for the experimental validation of the proposed RS control approach.}
    \centering
    \begin{tabular}{l|l}
    Parameter   & Value    \\ \hline
    $\Delta t$     & 0.001       \\
    $\alpha$ & 5 \\
    $r_\text{max}^a$   &  $2r_\text{min}^a$    \\
    $\kappa_r^a$,$\kappa_r^p$ & $30$ \\
    $\kappa_p^p$ & $2$ \\
    $r_\text{min}^p$   &  0.1   \\
    $r_\text{max}^p$   &  0.3   \\
    $\bm D_\text{track}$,$\bm K_\text{track}$  & $\text{diag}(40,40,40,40,40,40)$ \\
    $k_q$         & 250 \\
    $k_\text{sync}$         & 10 \\
    $w_{\text{track}}$,$w_{q}$,$w_{\text{sync}}$ & 1 \\
    $\Delta t_\text{int}$          & 0.1       
    \end{tabular}
    \label{tab:parameters}
    \end{table}
    
    To highlight the benefits of the proposed control framework, we perform an ablation study, comparing the proposed approach against three baseline approaches. The first baseline, referred to as the \emph{no RS} approach, is identical to the proposed approach aside from two key features proposed in the RS framework. The first removed feature is that the post-impact velocity reference is no longer adjusted based on the predicted post-impact velocity at impact time, implying that $\bm v^p_{o,d,f}(\bm p_o)$ in \eqref{eq:attractor} is now selected as the desired post-impact velocity reference field instead of $\bar{\bm v}_{o,d}^{p}(\bm p_o)$ from \eqref{eq:post_ext_vel}. The second removed feature is the interim mode, implying that the system switches directly from the ante-impact to the post-impact mode. This occurs upon detection of the impact for the hit-and-push case, and once an impact with both end effectors is detected for the dual-arm grabbing case. The reason for switching once both impacts are measured instead of after the first impact, is that this latter case would consistently result in a failed grasp. As the desired pose required a sideways motion of the end effectors, both end effectors would start moving in this direction before full contact is established, dropping the object and leading to an very unfavorable comparison. Due to the inability of state-of-the-art impact-aware control approaches like \cite{Yang2021} to deal with time-invariant references, this baseline represents the current state-of-the-art for time-invariant control under impacts.
    In the second and third baseline approaches, only one of the two aforementioned features is removed while retaining the other feature. The approach with \emph{no interim mode} switches directly from the ante-impact to post-impact mode, but does retain the adjustment of the post-impact velocity reference from the predicted impact map. Meanwhile, the approach with \emph{no impact map} does not retain this post-impact velocity reference adjustment, but does switch to the interim mode upon detection of the first impact. 
    
    Throughout the discussion of the experimental results, we will consider the target acceleration for the velocity tracking task as the main performance metric. This target acceleration is defined in \eqref{eq:a_a_t} for the ante-impact mode, in \eqref{eq:a_int_t} for the interim mode and in \eqref{eq:a_p_t} for the post-impact mode. Naive control strategies can lead to peaks in the velocity tracking error at the time of impact, which will then appear in the target acceleration, and as a result as undesired peaks in the input torques $\bm \tau_i^*$ computed through \eqref{eq:EOM_free}. By quantifying the effect on target acceleration, we gain insight in the severity of the unwanted input peaks, and we can relate these peaks directly to the corresponding velocity tracking errors.
    
    In the remainder of this section, we will first focus on the simpler hit-and-push experiments, followed by the more challenging dual-arm grabbing experiments.

    \subsection{Hit-and-push experiments}    

\begin{figure}
		\centering
		\includegraphics[width=\linewidth]{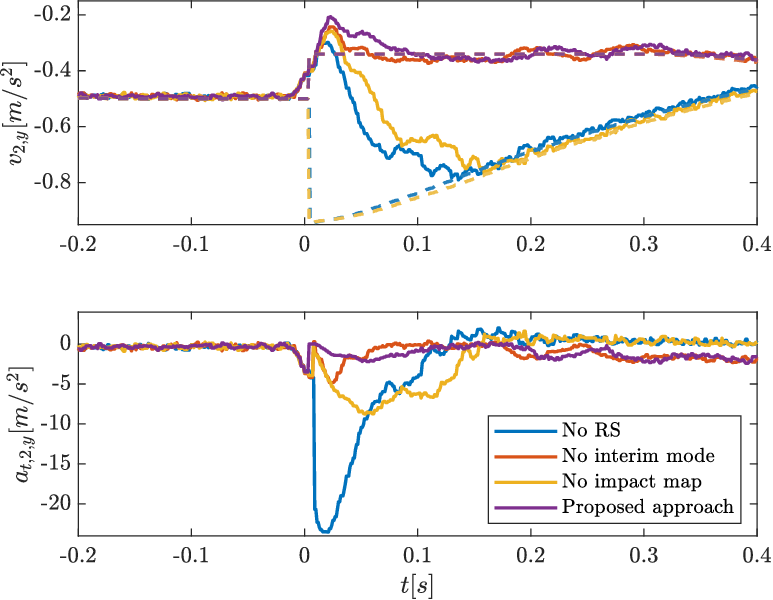}
		\caption{Velocity and target acceleration for a set of hit-and-push experiments with the juice pack as depicted in Figure \ref{fig:snapshots_push}. In blue, red and yellow, the results for the three baseline control approaches are depicted and, in purple, the results for the proposed time-invariant RS approach are shown. The dashed line in the top plot represents the velocity reference that corresponds to the control mode at that time. The experiments are aligned in time such that the impact detection always occurs at $t =$ 0s.}
		\label{fig:Vel_acc_plots_push}
	\end{figure}

    As shown in Figure \ref{fig:snapshots_push}, the proposed time-invariant RS approach can be used to successfully execute a hit-and-push motion. To highlight the effect of the proposed approach around the impact event, Figure \ref{fig:Vel_acc_plots_push} shows the linear velocity and corresponding velocity reference (top) and the corresponding target acceleration (bottom) for a set of four experiments. These experiments all have identical conditions as seen in Figure \ref{fig:snapshots_push}, but employ either one of the three baseline control approaches, or the proposed RS approach. Only the velocity in the $y$-direction is shown, because the $y_A$-axis illustrated in Figure \ref{fig:setup} is aligned with the impact direction. For easy comparison, the impact timings in the experiments are aligned in these plots to show the impact always occurring at $t = $ 0s. 
       
    \begin{figure}
		\centering
		\begin{subfigure}[b]{\linewidth}
			\centering
			\includegraphics[width=0.97\textwidth]{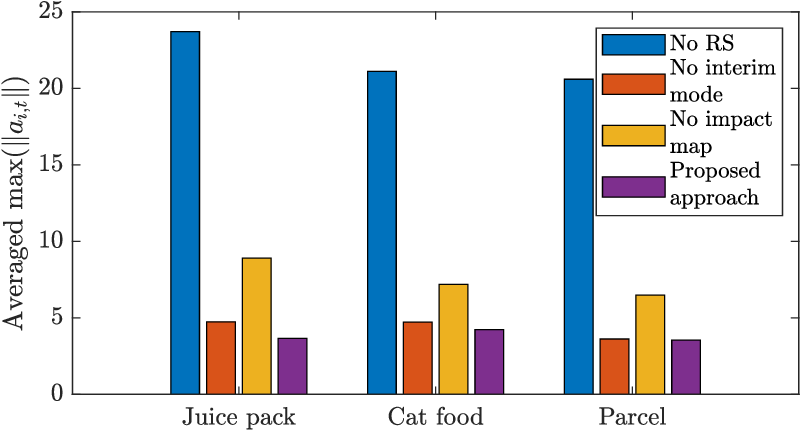}
			\caption{Different objects}
			\label{fig:Diagram_object_push}
		\end{subfigure}
		\hfill
		\begin{subfigure}[b]{\linewidth}
			\centering
			\includegraphics[width=0.97\textwidth]{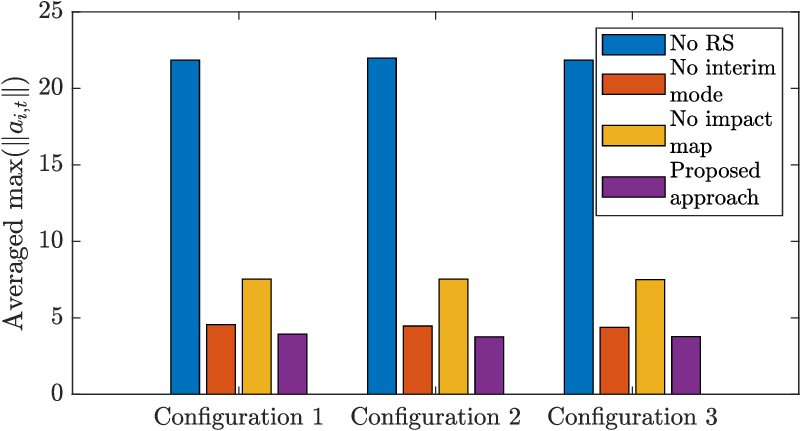}
			\caption{Varying initial configurations}
			\label{fig:Diagram_init_push}
		\end{subfigure}
		\hfill
		\begin{subfigure}[b]{\linewidth}
			\centering
			\includegraphics[width=0.97\textwidth]{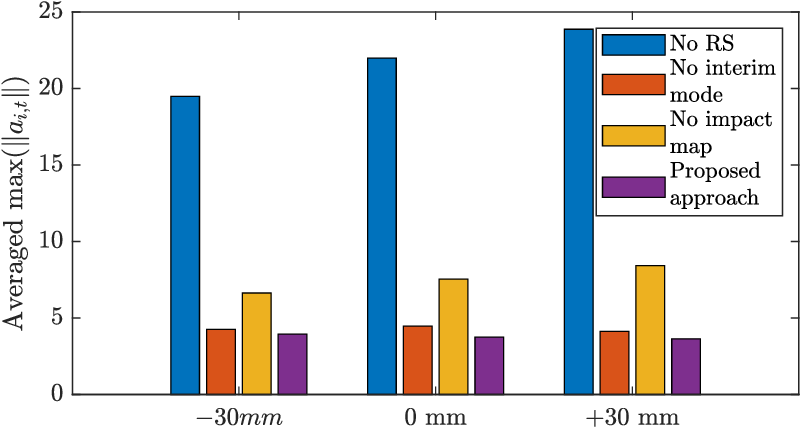}
			\caption{Displacement of the object in $y_A$-direction}
			\label{fig:Diagram_disp_push}
		\end{subfigure}
        \caption{Diagrams showing the averaged maximum target acceleration norms for the hit-and-push experiments. Each bar represents the average target acceleration for all experiments with a given controller (indicated by the color) and a given object/initial configuration/initial object displacement.}
		\label{fig:diagram_push}
	\end{figure}
 
    In the velocity plot of Figure \ref{fig:Vel_acc_plots_push}, it can be seen that for all experiments, the velocity initially jumps from the commanded $-0.5$ m/s towards the predicted post-impact velocity of $-0.34$ m/s, which is the velocity reference for the proposed approach and the approach with no interim mode. Both of the other two baseline approaches do not take the impact map into consideration in the formulation of the post-impact velocity reference, causing a large velocity tracking error as soon as the impact is detected. This is reflected in the target acceleration shown in the bottom plot, where the approach with no RS (blue line) immediately enters the post-impact mode, and thus immediately tracks this velocity reference with relatively high gains, causing an enormous jump in the target acceleration that is clearly unwanted, as it can induce hardware damage and cause unpredictable and potentially even unstable behaviour. Due to the initial removal of velocity feedback and slow increase of the velocity feedback gain in the interim mode, this temporary increase of the norm of the feedback is less sudden in the approach with no impact map (yellow line), also causing a slower convergence to the post-impact velocity. However, the increase in feedback action is still significant compared to the input signal well before and after the impact event (e.g. at $-0.2$s and $0.4$s). While the reference velocity jump is admittedly very large and could already be reduced by manually tuning the post-impact velocity reference, the lack of such peaks for the proposed approach shows that the need for such a cumbersome manual tuning procedure is removed when the impact map is built into the reference. 
    
    Comparing the proposed approach (purple line) with the approach with no interim mode (red line), the need to include the interim mode for this use case is less visible immediately, assuming a proper post-impact reference is selected. A peak in the target acceleration can be spotted after the impact is detected for the approach with no interim mode, which is caused by a moderate peak in the tracking error while the velocity is converging to the final post-impact reference. But this peak is significantly smaller than that of the other two baselines, and is roughly as high as the acceleration peak in each of the experiments prior to impact detection, which is caused by a small delay in the impact detection procedure. Thus, this peak is likely to have less of an effect on the performance. One caveat is that the size of the input peak could definitely increase if the control gains are further increased. But since a higher gain than the currently selected value would result in failed experiments for the baseline approaches, especially the no RS approach, the values in Table \ref{tab:parameters} are used everywhere to allow for a fair comparison.

    To confirm the claim that the time-invariant nature of the control approach suits a large range of initial conditions, experiments are performed for three significantly different initial conditions. Furthermore, to confirm the claim that the control approach is robust against uncertainty in the impact timing/location, experiments are performed with different object placements without adjusting the velocity references. For the hit-and-push experiments, experiments are performed with the object in its predicted location, as well as a displacement in $y_A$-direction indicated in Figure \ref{fig:setup} of -30 mm and +30 mm, causing a delayed or early impact, respectively. A total of 5 experiments is performed per combination of physical setup and control approach, leading to a total of 300 pushing experiments that have been performed. In each of these experiments, the maximum norm of the target acceleration is recorded. We then averaged these maximum norms for the different control approaches between different groups to investigate the effect of changes in the physical setting on the controller behaviour. This results in a diagram used to investigate the effect of taking a different object in Figure \ref{fig:Diagram_object_push}, different initial conditions in Figure \ref{fig:Diagram_init_push}, and the object displacement in Figure \ref{fig:Diagram_disp_push}. 

    Starting off with the influence of the different objects in Figure \ref{fig:Diagram_object_push}, it is noticeable that the approach with no RS and the approach with no impact map show the largest target acceleration peak for the object with the largest inertia (the juice pack) and the smallest peak for the object with the smallest inertia (the parcel). This is caused by the magnitude of the velocity jump; a larger inertia implies that the post-impact velocity will be smaller, and will thus be further removed from the post-impact velocity references for these baseline approaches. The proposed approach is not affected by this because the predicted velocity jump is updated based on the object, so the differences between the velocity jumps are embedded in the references. The approach with no interim mode does seem to be affected more by the object selection, which is caused by an increase of impact-induced vibrations when objects with higher inertia are impacted. This effect is also not seen in the proposed control approach due to the interim mode being designed to be less affected by vibrations that occur immediately after the impact. 

    \begin{figure*}
		\centering
		\includegraphics[width=0.93\linewidth]{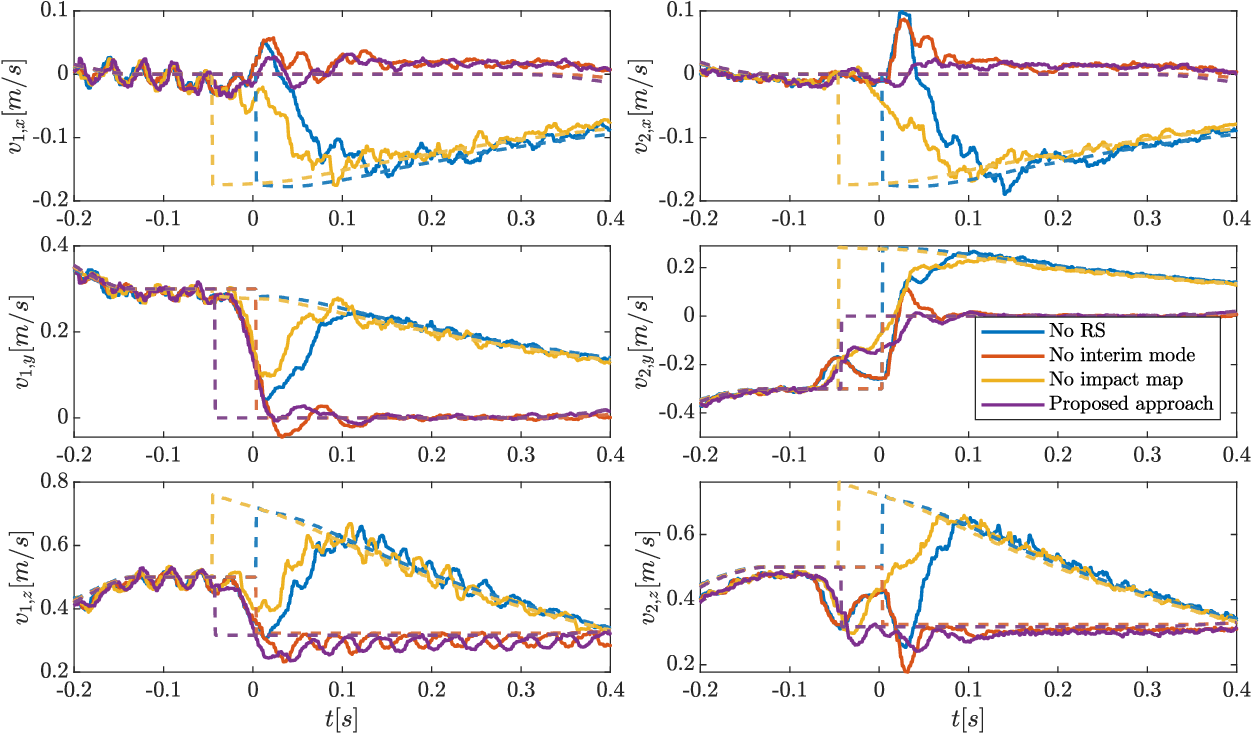}
		\caption{Velocity for a set of dual-arm grabbing experiments with the box of cat food as depicted in Figure \ref{fig:snapshots_grab}. In blue, red and yellow, the results for the three baseline control approaches are depicted and, in purple, the results for the proposed time-invariant RS approach are shown. The dashed line represents the velocity reference that corresponds to the control mode at that time. The left plots show the velocity of the left robot (robot 1), and the right plots show the velocity of the right robot (robot 2). The experiments are aligned in time such that the impact is detected for both robots at $t =$ 0s.}
		\label{fig:Vel_plots_grab}
	\end{figure*} 
    \begin{figure*}
		\centering
		\includegraphics[width=0.93\linewidth]{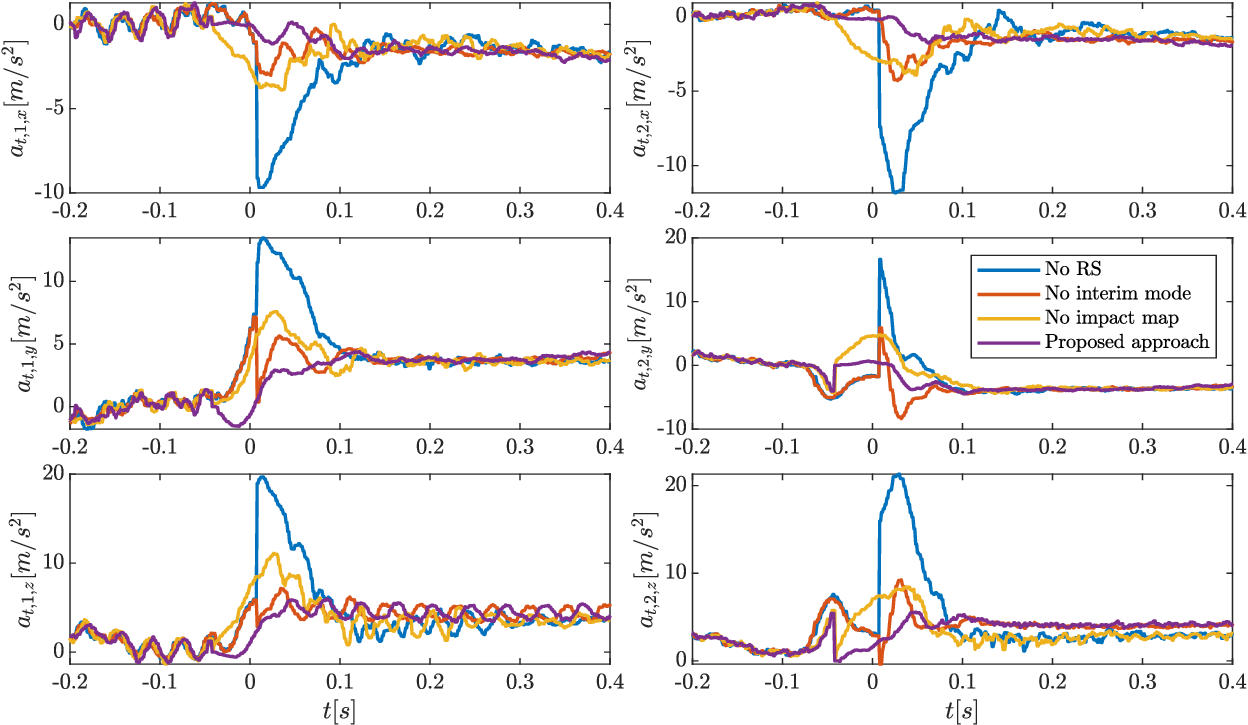}
		\caption{Target acceleration for a set of dual-arm grabbing experiments with the box of cat food, corresponding to the velocity plots shown in Figure \ref{fig:Vel_plots_grab}.}
		\label{fig:Acc_plots_grab}
	\end{figure*}

    The influence of the initial configuration, shown in Figure \ref{fig:Diagram_init_push}, is essentially nonexistent for each of the control approaches, as the ante-impact mode  will for each controller cause a similar impact state, with the small difference in this impact state apparently being negligible. This is a direct benefit of the use of time-invariant references, which are by design robust to different initial configurations. Note that also here the average peak acceleration is the lowest for the proposed approach.

    Finally, the influence of the object displacement does seem to affect the approach with no RS and the approach with no impact map significantly, while no effect can be seen for the proposed approach and the baseline approach with no interim mode. This occurs because the velocity field $\bm v^p_{o,d,f}(\bm p_o)$ from \eqref{eq:attractor} commands a higher velocity if the object is placed further away from the desired final object location $\bm p_{o,f}$, as is the case for the +30 mm displacement case. Since this velocity field is not yet followed near the impact in the proposed approach and the approach with no interim mode, these approaches remain practically unaffected.

    In conclusion, for all of the scenarios considered in Figure \ref{fig:diagram_push}, the proposed approach results in the lowest averaged maximum desired acceleration norm, highlighting the effectiveness of the proposed control approach in reducing input peaks for the hit-and-push scenario.

    \subsection{Dual-arm grabbing experiments}

    The same analysis as for the hit-and-push use case has been performed for the dual-arm grabbing experiments, illustrated in Figure \ref{fig:snapshots_grab}. Figure \ref{fig:Vel_plots_grab} shows the velocities for experiments with an identical physical setting and different controllers (3 baseline approaches and the proposed RS approach). The signals are aligned in such a way that detection of the second impact always occurs at $t=$ 0s, implying that contact on both end effectors is made at this time. As can be seen in Figure \ref{fig:snapshots_grab}, the object is displaced with respect to its estimated position, causing the right robot (robot 2) to impact the object before the left robot (robot 1), which is confirmed in Figure \ref{fig:Vel_plots_grab} as the velocity jumps occur earlier for the robot 2. 
    
    Considering the approach with no RS and the approach with no impact map, we observe the same pattern as for the hit-and-push use case, as the velocity reference jumps away from the actual velocity signal, causing a large velocity tracking error for both robots in all directions, and a large corresponding target acceleration, as shown in Figure \ref{fig:Acc_plots_grab}. Again, the interim mode reduces, but does not remove this effect for the approach with no impact map. 
    
    However, unlike in the hit-and-push experiments, we do see a larger distinction between the results for the proposed approach and the approach with no interim mode. First, we see that the velocity reference in the proposed approach jumps before the reference in the approach with no interim mode. This is because in the proposed approach, the interim mode is initiated upon detection of the first impact, while the approach with no interim mode jumps directly from the ante-impact mode to the post-impact mode upon detection of the second impact. The approach with no interim mode thus remains in the ante-impact mode while the velocity has already started to jump, causing a velocity error and a corresponding target acceleration peak before $t=$ 0s. A moderate peak in the target acceleration is also visible for the proposed approach due to a delay in impact detection, but, as soon as the first impact is detected, the target acceleration reduces to a value near zero for all directions. Furthermore, the effect of impact-induced vibrations on the velocity tracking error is increased compared to the hit-and-push case due to the heavier impact. Because of this fact, the target acceleration in the approach with no interim mode fluctuates and peaks more heavily after the second impact is detected at $t=$ 0s. However, for the proposed approach, these input peaks are essentially removed due to the interim mode design.

    \begin{figure}
		\centering
		\begin{subfigure}[b]{\linewidth}
			\centering
			\includegraphics[width=\textwidth]{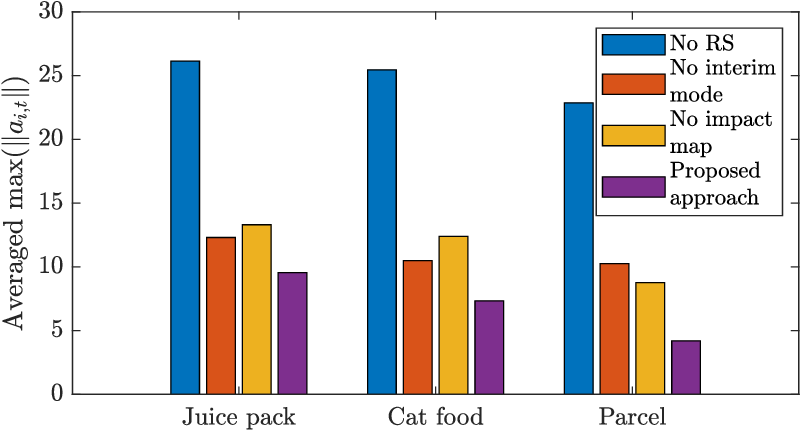}
			\caption{Different objects}
			\label{fig:Diagram_object_grab}
		\end{subfigure}
		\hfill
		\begin{subfigure}[b]{\linewidth}
			\centering
			\includegraphics[width=\textwidth]{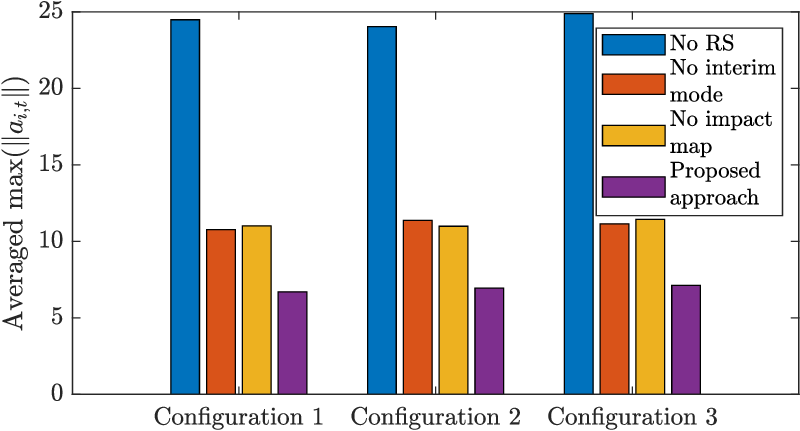}
			\caption{Varying initial configurations}
			\label{fig:Diagram_init_grab}
		\end{subfigure}
		\hfill
		\begin{subfigure}[b]{\linewidth}
			\centering
			\includegraphics[width=\textwidth]{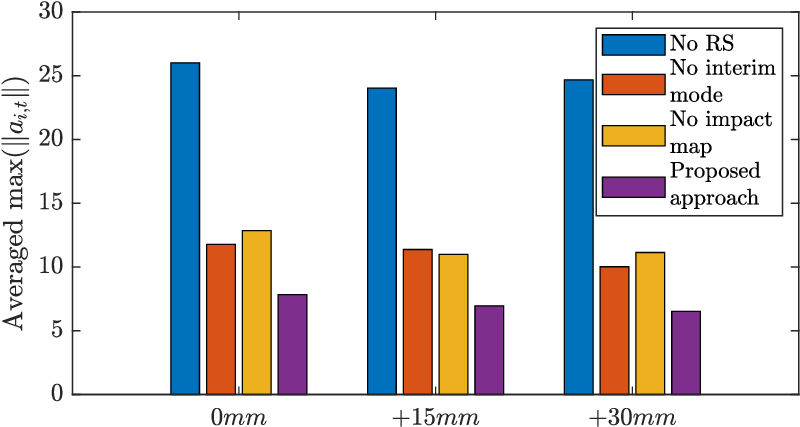}
			\caption{Displacement of the object in $y_A$-direction}
			\label{fig:Diagram_disp_grab}
		\end{subfigure}
        \caption{Diagrams showing the averaged maximum target acceleration norms for the dual-arm grabbing experiments. Each bar represents the average target acceleration for all experiments with a given controller (indicated by the color) and a given object/initial configuration/initial object displacement.}
		\label{fig:Diagram_grab}
	\end{figure}

    Similar to the hit-and-push case, experiments are performed for three significantly different initial conditions. For the dual-arm grabbing case, these initial conditions also vary significantly in the initial symmetry between the two robots with respect to the object, showcasing the influence of the synchronization task in the ante-impact mode. 
    Experiments are also performed with different object placements without adjusting the velocity references. For the grabbing experiments, this is evaluated for a 0mm, a +15mm and a +30mm displacement in $y_A-$direction of the object with respect to its estimated position, causing the right robot to impact the object increasingly earlier than the left robot. The maximum target acceleration norms are then and computed and averaged over both robots and all experiments in these different groups. The results for these 300 experiments are shown in Figure \ref{fig:Diagram_grab}.

    Starting with the different objects compared in Figure \ref{fig:Diagram_object_grab}, we see that the average maximum target acceleration norm decreases with a reduced inertia of the object for all control approaches, including the proposed approach. This is explained by the influence of the feedforward target wrench $\vf^p_{i,t}$ in \eqref{eq:a_p_t}. This target wrench is tasked with clamping the object, and a higher inertia of the object implies that a higher clamping force is required to prevent slippage, in turn increasing the target acceleration. Aside from this fact, we see that the proposed approach results in the lowest maximum target acceleration for all of the objects, as expected. 
    
    For the different initial configurations, compared in Figure \ref{fig:Diagram_init_grab}, we can observe a minor effect on the maximum target acceleration for the baseline approaches. This effect is most likely caused by a slightly different impact configuration, resulting in a different velocity jump between different configurations. The effect appears to be reduced for the proposed approach, likely because the predicted velocity jump in the reference is updated based on the configuration of the system at the impact time. 
    
    For the different displacements, compared in Figure \ref{fig:Diagram_disp_grab}, we observe no noticeable consistent effect on the target acceleration upon increasing the initial displacement for either control approach. One might intuitively expect larger displacements result in an increased target acceleration  for the approaches with no interim mode and no RS, because the interim mode is designed to add robustness to a loss of simultaneity of the impacts, which occurs when the objects are displaced. However, while increasing the displacement does increase the time duration where the target acceleration is increased, the maximum target acceleration appears to occur at the time of switching to the post-impact mode due to the aforementioned fluctuations in the velocity directly after the second impact. Meanwhile, the low target acceleration for the proposed approach does prove its robustness to uncertainty in the object location and a loss of impact simultaneity.

    Finally, we can conclude for the dual-arm grabbing experiments that, as with the hit-and-push experiments, the proposed approach results in the lowest averaged maximum desired acceleration norm as shown in Figure \ref{fig:diagram_push}, highlighting the effectiveness of the proposed control approach in reducing input peaks.

    \section{Conclusion}
    \label{sec:conclusion}
    
    In this work, we have presented an impact-aware control approach that prevents peaks in the input signals as a result of impact-induced velocity jumps. 
    The approach uses the time-invariant reference spreading control framework, which creates ante-impact and post-impact velocity fields that are coupled via an impact map to reduce the velocity tracking error when switching to the post-impact control mode. This impact map is extracted from a series of impact simulations using a nonsmooth physics engine, which is known to be reliable through a validation performed in \cite{Steen2024a}. 
    Aside from an ante-impact and a post-impact control mode, a novel interim mode design is presented. This interim mode uses the ante-impact velocity reference field to generate a position feedback signal that promotes contact completion, and combines the idea with a heuristic blending approach that ensures a smooth transition between the different control modes. 
    By the grace of the time-invariant nature of the control approach, the control approach supports a large range of initial conditions without requiring replanning of the reference, and will not unnecessarily catch up to a time-based reference in the presence of disturbances. 

    This work also presents, for the first time, an experimental validation of the time-invariant reference spreading control framework. This extensive validation based on 600 experiments for two different use cases and different objects shows that the proposed approach consistently prevents peaks in the input signals compared to three different baseline control approaches. 

    One of the aspects for future research on (time-invariant) reference spreading would be the investigation into a detection scheme to detect contact completion, with the aim of further improving the interim mode. With its current design, the interim mode lasts for a user-defined time, which can be problematic in cases of unknown levels of uncertainty in the environment, and may require manual tuning for different types of object. An interim mode based on contact completion detection could reduce this issue by only switching to the post-impact mode when there is certainty that the contact event has been completed. 
    Another interesting topic for future research would be the application and validation of this control framework to more complex robots, such as full humanoids, which is already enabled in this work by casting the controller into a QP control framework, suitable for dealing with multiple (conflicting) tasks and constraints as required for humanoid control. Finally, from a theoretical perspective, the proposed control approach would benefit from an investigation into a formal proof of stability as a formal certificate of its effectiveness. 
    
	\addtolength{\textheight}{-0cm}  
	
	\section*{APPENDIX}

	\subsection{Post-impact reference RBF interpolation}\label{sec:interpolation}

    Here, we will give details on the procedure to extract the post-impact predicted velocity $\bm v_{o,\text{est}}^+$, which is used in the formulation of the post-impact velocity reference, from a set of impact simulations using a nonsmooth physics engine. 
    To this end, the impact state is registered when an impact is first detected, and the end effector position is projected onto a plane parallel to the nominal impact surface to obtain the projected end effector position $\bm p^-_{p,i} \in \mathbb{R}^2$. This position is then compared to the projected end effector positions $\bm p^-_{p,i,j}$ in each simulation $j \in \{1, \dots, N_\text{exp}\}$ through radial basis function (RBF) interpolation.     
    %Assuming a single simultaneous impact with sticking contact between the end effector and the object, the post-impact object velocity $\bm v_{o,j}^+$ is estimated as $\bm v_{o,j}^+ = \frac{1}{2} \left(\bm v_{1,j}^+ + \bm v_{2,j}^+\right)$. 
    As such, a weighting matrix $\bm W = (\bm w_1, \bm w_2, \dots , \bm w_{N_\text{exp}}) \in \mathbb{R}^{3 \times N_\text{exp}}$ is determined through
    \begin{equation}
        \bm W = \bm \Phi^{-1} \begin{bmatrix}
            \bm v_{o,1}^+ & \bm v_{o,2}^+ & \dots & \bm v^+_{o,N_\text{exp}}
        \end{bmatrix},
    \end{equation}
    where     
    $$\resizebox{\linewidth}{!}{\arraycolsep=2pt%
    $\bm \Phi=\left[\begin{array}{cccc}
    \phi\left(\left\|\bm p^-_{p,i,1}-\bm p^-_{p,i,1}\right\|\right) & \cdots & \phi\left(\left\|\bm p^-_{p,i,1}-\bm p^-_{p,i,{N_\text{exp}}}\right\|\right) \\
    \vdots & \ddots & \vdots \\
    \phi\left(\left\|\bm p^-_{p,i,{N_\text{exp}}}-\bm p^-_{p,i,1}\right\|\right) & \cdots & \phi\left(\left\|\bm p^-_{p,i,{N_\text{exp}}}-\bm p^-_{p,i,{N_\text{exp}}}\right\|\right)\end{array}\right].$}$$    
    The exponential radial basis function $\phi:\mathbb{R}\to\mathbb{R}$ is given by $$\phi(r)=e^{-(\rho r)^{2}},$$
    with user-defined shaping parameter $\rho \in \mathbb{R}^+$. The interpolated predicted post-impact box velocity is finally obtained as 
    \begin{equation}\label{eq:v_p_est}
        \bm v_{o,\text{est}}^+ = \sum_{j=1}^{N_\text{exp}} \bm w_{j} \phi\left(\left\|\bm p_{p,i}^--\bm{p}^-_{p,i,j}\right\|\right).
    \end{equation}
    Note that this formulation of $\bm \Phi$ and $\bm v_{o,\text{est}}^+$ holds specifically for a single-arm use case. For dual-arm grabbing, we replace $\bm p^-_{p,i}$ and $\bm p^-_{p,i,j}$ by $[\bm p^-_{p,1}; \bm p^-_{p,2}]$ and $[\bm p^-_{p,1,j}; \bm p^-_{p,2,j}]$, respectively.
    %with $\bm p^- = [\bm p_{1,x}^-; \bm p_{1,z}^-; \bm p_{2,x}^-; \bm p_{2,z}^-]$ as the measured end effector positions at the time of impact in the direction tangential to the desired impact surface.

\bibliography{References/library}{}
\bibliographystyle{ieeetr}

\end{document}